\documentclass[preprint,12pt]{elsarticle}

\usepackage{amsmath,amsfonts}
\usepackage{array}
\usepackage[caption=false,font=normalsize,labelfont=sf,textfont=sf]{subfig}
\usepackage{textcomp}
\usepackage{stfloats}
\usepackage{url}
\usepackage{verbatim}
\usepackage{graphicx}
\usepackage[small]{caption}
\usepackage{arydshln}
\usepackage{amssymb}
\usepackage[inkscapearea=page]{svg}
\usepackage{adjustbox}
\usepackage{multirow}
\usepackage{fancyhdr,graphicx}
\usepackage{hyperref}
\usepackage{fancyhdr,graphicx,amsmath,amssymb}
\usepackage[normalem]{ulem}

\usepackage{float}
\usepackage[ruled,vlined,commentsnumbered]{algorithm2e}
\usepackage[noend]{algpseudocode}

\usepackage{xurl}

\usepackage{fancyhdr}
\usepackage{lipsum} 
\usepackage{geometry}
\usepackage{etoolbox}

\fancypagestyle{firstpagefooter}{
    \fancyhf{} 
    \fancyfoot[L]{\footnotesize This work has been submitted to the IEEE for possible publication. Copyright may be transferred without notice, after which this version may no longer be accessible.}

}

\biboptions{sort&compress}

\makeatletter
\def\ps@pprintTitle{%
 \let\@oddhead\@empty
 \let\@evenhead\@empty
 \let\@oddfoot\@empty
 \let\@evenfoot\@empty
}
\makeatother

\begin{document}
\thispagestyle{firstpagefooter}

\begin{frontmatter}



\title{
Anomaly Detection for Non-stationary Time Series using Recurrent Wavelet Probabilistic Neural Network
}


\author[1]{Pu Yang}
\author[1]{J.~A.~Barria\corref{cor1}}
\cortext[cor1]{Corresponding author}
\ead{j.barria@imperial.ac.uk}


\affiliation[1]{organization={Department of Electrical and Electronic Engineering, Imperial College London}, 
                addressline={South Kensington Campus}, 
                city={London}, 
                postcode={SW7 2AZ}, 
                country={United Kingdom}}

\fntext[label1]{Tel: +44 (0)20 7594 6275}

\begin{abstract}
In this paper, an unsupervised Recurrent Wavelet Probabilistic Neural Network (RWPNN) is proposed, which aims at detecting anomalies in non-stationary environments by modelling the temporal features using a nonparametric density estimation network. The novel framework consists of two components, a Stacked Recurrent Encoder-Decoder (\textit{SREnc-Dec}) module that captures temporal features in a latent space, and a Multi-Receptive-field Wavelet Probabilistic Network (\textit{MRWPN}) that creates an ensemble probabilistic model to characterise the latent space. This formulation extends the standard wavelet probabilistic networks to wavelet deep probabilistic networks, which can handle higher data dimensionality. The \textit{MRWPN} module can adapt to different rates of data variation in different datasets without imposing strong distribution assumptions, resulting in a more robust and accurate detection for Time Series Anomaly Detection (TSAD) tasks in the non-stationary environment. We carry out the assessment on 45 real-world time series datasets from various domains, verify the performance of RWPNN in TSAD tasks with several constraints, and show its ability to provide early warnings for anomalous events.
\begin{center}
{\tiny \textit{This work has been submitted to the IEEE for possible publication. Copyright may be transferred without notice, after which this version may no longer be accessible.}}
\end{center}
\vspace{0.5em}
\end{abstract}




\end{frontmatter}




\section{Introduction}
Anomaly detection (AD), which is a subclass of the classification task, aims at detecting abnormal patterns that deviate from the regular data distribution \cite{Atkinson1981, Aggarwal2013}. In industry, AD has been widely used in different fields, including network intrusion detection and prediction, credit card transaction verification, medical diagnosis, and environment changes \cite{Alarcon-Aquino2001AnomalyWavelets, Alarcon-Aquino2006MultiresolutionPrediction, Aggarwal2013, Wang2019a, Chalapathy2019}.

Among the variety of data collected from real-time applications, e.g., image, video stream, or text, time series data has been one of the increased interests. Related applications include AD in communication networks, network traffic analysis, and Internet-of-Things (IoT) devices \cite{Alarcon-Aquino2001AnomalyWavelets, Ren2019Time-seriesMicrosoft, Liu2021DeepApproach, Vu2022LearningDetection}.
Some underlying challenges when analysing real-time services include, (i) potential restrictions in the amount of data available for model training, due to privacy regulations, the infrequent occurrence of anomalous data samples in AD tasks, or the high cost of data collection and labelling;
(ii) the prediction model needs to be able to deal with unseen data, which might exhibit different statistical properties from previously observed.

Existing AD solutions can be summarised into three categories \cite{Braei2020AnomalyState-of-the-Art, Wen2023TransformersSurvey}: the first category, known as statistical approaches, includes examples such as the Autoregressive Model and Autoregressive Integrated Moving Average. The second group comprises the general machine learning approaches, which include local outlier factors, One-Class Support Vector Machines, and Isolation forests. The last category includes the deep-learning-based (DL) approaches, where Multi-layer Perceptrons (MLPs), Convolutional Neural Networks (CNNs), Recurrent Neural Networks (RNNs), 
and Transformers 
are commonly used to learn the expressive representations of the source data, which can then be used for AD. 

Recurrent neural networks (RNNs) and their variants, such as long short-term memory (LSTM) and gated recurrent unit (GRU), are commonly used for the classification and prediction of time series data. The hidden states generated by the RNNs represent compressed features of the original sequence, which can then be used for future data prediction and classification. One way to achieve this is through the sequence-to-sequence (\textit{Seq2seq}) structure, which has been proposed and used in text generation and machine translation \cite{Sutskever2014SequenceNetworks}. The encoder-decoder framework in \textit{Seq2seq}, has also been widely applied in time series anomaly detection (TSAD) tasks \cite{Malhotra2015LongSeries, Malhotra2016LSTM-basedDetection, Guo2018MultidimensionalApproach}.

A common unsupervised encoder-decoder-based AD approach is to train a discriminative model on the majority class by mapping the original data with latent features; the anomalies can then be detected based on the deviation from the normal behaviour. For instance, 
\cite{Malhotra2016LSTM-basedDetection} utilised the hidden states of RNNs to reconstruct the original data from the compressed features and measure the reconstruction error as an anomaly score. In \cite{Xue2024}, sliding windows were used to extract multi-scale features for TSAD. In \cite{Yao2024}, two encoders were proposed to capture features in temporal and variable dimensions, while Fourier Transform was used to preserve frequency-domain characteristics and prevent over-smoothing during the reconstruction process.

Another unsupervised approach is to build a probabilistic model by modelling the distribution of the latent space. Examples such as, \cite{Zong2018DeepDetection} introduced an autoencoder (AE)-based model that leverages the latent features from the encoder and adds distance metrics to enhance the reconstruction process. The compressed features are then fed to a Gaussian mixture model (GMM) to estimate the sample energy for anomaly detection. Moreover, 
the method proposed in \cite{Guo2018MultidimensionalApproach} employs a GRU-based variational autoencoder (VAE) for data compression and reconstruction, and a GMM is used to model the latent space. 
In \cite{Wang2022VariationalSeries}, a Transformer-based VAE was introduced to extract multi-scale information for the VAE to detect the anomalies. 

Existing probabilistic solutions rely on the parametric density estimation techniques, as the distribution parameters (i.e., mean and standard deviation) can be jointly optimised with the model training process. While the parametric-based methods impose strong distribution assumptions on the target data, the nonparametric-based density estimation techniques, such as kernel density estimators and wavelet density estimators (WDEs), have minimal assumptions on the data distribution \cite{alma995345184401591, Vidakovic1999StatisticalStatistics}.  

Wavelet Density Estimators (WDEs), 
leverages
the advantage of wavelet multiresolution analysis, and has been shown to be more effective than other nonparametric methods as it can capture the transient phenomena (e.g., density discontinuities) and non-stationary characteristics \cite{Vidakovic1999StatisticalStatistics, GARCIATREVINO2019111}. Wavelet (multiresolution) analysis can capture time and frequency information at the same time by decomposing the data into different scales and is also known to be robust with sparsity in compression, denoising, and density estimation \cite{Mallat2009}. Hence, the analysis based on wavelet multiresolution enables the detection and investigation of transient phenomena and signals with non-stationary characteristics, which are common in real-world time series data (see examples in \cite{Alarcon-Aquino2006MultiresolutionPrediction, GARCIATREVINO2019111}). 

A standard WDE has two limitations that hinder its further research with deep neural networks (DNNs): (i) the lack of an analytical form which leads to high computational complexity, and (ii) it suffers from the curse of data dimensionality. In \cite{GARCIATREVINO2019111}, a novel density estimator called Radial Wavelet Frame Density Estimator is proposed to overcome the first issue and handle higher data dimensionality. It provides a low and constant computational complexity while maintaining the benefits of WDEs. Moreover, we proposed a wavelet probabilistic neural network in \cite{Garcia-Trevino2024WaveletNetworks}, which aims at data stream classification in the non-stationary environment, where the data length is unconstrained, and concept drift may occur over time.

We address the limitations of the existing discriminative and generative-based TSAD methods, which can be summarised into the following three points: (i) The anomaly score is based on the reconstruction or prediction process. For the non-stationary data, such scores can no longer be used to correctly describe the class differences. (ii) Existing generative-based methods build parametric models to describe the latent space, which impose strong data distribution assumptions and such parametric models cannot handle non-stationary environments. Moreover, they become less effective when densities with discontinuous and localised features exist. (iii) Deep neural networks usually have higher requirements on the amount of training data, making them very difficult to learn a generalised model with limited training data available.

The proposed unsupervised framework, called Recurrent Wavelet Probabilistic Neural Network (RWPNN), aims at detecting anomalies in the non-stationary environment by modelling the latent space using a WDE-based probabilistic network.
To the best of our knowledge, RWPNN is the first approach to formulating a WDE in an unsupervised deep neural network, overcoming the high computational cost and the curse of dimensionality that exist in the current solutions. 


RWPNN is particularly useful for imbalanced training data with scarcity in the non-stationary environment. The framework comprises two modules: a stacked Recurrent Encoder-Decoder (\textit{SREnc-Dec}) module that extracts the temporal features using RNNs; and a novel Multi-receptive-field wavelet probabilistic network (\textit{MRWPN}) module that offers an ensemble view of different rates of data variation in the latent space by creating an ensemble probabilistic model composed of multiple probability density functions that characterise the latent space. By analysing the probabilistic information generated from the \textit{MRWPN}, the onset of deviation from the normal patterns can be identified, which can then be used in an early warning system.
Note that by combining the \textit{SREnc-Dec} with the \textit{MRWPN}, we extend the application of the current generative-based wavelet probabilistic network from standard neural networks to deep neural networks.

Two are the main contributions of RWPNN. (i) The first one is modelling the latent space with suspected concept drifts by formulating a nonparametric WDE-based deep neural network. Note that the WDE-based probabilistic network for modelling the latent space with concept drifts in an Encoder-Decoder network has not been investigated before due to the lack of an efficient wavelet-based solution. (ii) Existing wavelet-based probabilistic networks can only monitor a single rate of data variation; the \textit{MRWPN} module, on the other hand, can capture different rates of variation simultaneously without formulating extra networks and hence, not compromising the time complexity.

Note that existing solutions we discussed, such as CNN \cite{Liu2021DeepApproach}, RNN \cite{Malhotra2015LongSeries, Malhotra2016LSTM-basedDetection}, GMM and VAE \cite{Zong2018DeepDetection, Guo2018MultidimensionalApproach}, cover different neural network architectures for TSAD. Yet, few of them consider the paucity of training data and target the non-stationary environment. These methods detect anomalies by modelling the data reconstruction, data prediction, and modelling the latent space using parametric methods like Gaussian Mixture Models; and therefore, their performance will be affected in the non-stationary environment. 

RWPNN, on the other hand, opens a new venue of TSAD in a non-stationary environment by modelling the latent space in a probabilistic term using a nonparametric wavelet probabilistic network (\textit{MRWPN}), which provides an ensemble view that characterises the space.  
Moreover, further research of the \textit{MRWPN} module will not be affected by the curse of dimensionality, as a smaller but more informative latent space can be formed by leveraging the \textit{SREnc-Dec} module and therefore, RWPNN can adapt to time series data with higher dimensionality.

The rest of the paper is organised as follows. In Section \ref{sec:background}, the background of the proposed RWPNN is briefly reviewed. Section \ref{sec:wpnn} reviews the existing wavelet probabilistic network. The RWPNN formulation is given in Section \ref{sec:method}. The performance comparisons between RWPNN and benchmark algorithms are assessed in Section \ref{sec:result}. Finally, the conclusions of the work are presented in Section \ref{sec:final_remark}.

\section{Related research}\label{sec:background}
\subsection{Recurrent Neural Networks}
Recurrent Neural Networks (RNNs) are designed for capturing temporal dependencies from sequential data, which is widely used in machine translation, stock market prediction, and TSAD \cite{Choi2021DeepGuidelines}. The key feature in RNN is the hidden state $h_t$ at the timestamp $t$; it captures and retains the compressed information of the input sequence, which can be used for the next available data at $t+1$. Consider a univariate time series sequence with unconstrained length $\mathbf{x} = [x_1, x_2, ..., x_{\infty}]$ and a sliding window with size $L$, the sequence $\mathbf{x} = [x_1, x_2, ..., x_{L}]$ will be used as the input of RNN; at timestamp $t$, the corresponding temporal dependencies of this sequence will be stored in the hidden state $h_t$.

Conventional RNNs suffer from the vanishing gradient problem due to longer sequence lengths. To solve the problem, Long Short-Term Memory (LSTM) and Gated Recurrent Unit (GRU) have been proposed. The formulation of LSTM is given in Equation \ref{eq:LSTM}, where $\sigma$ is the non-linear activation function \textit{Sigmoid}, $\otimes$ is the element-wise product operation, and $\oplus$ is the element-wise addition operation. Two hidden states $C_t$ and $h_t$ are provided, with $C_t$ being the long-term memory, and $h_t$ being the short-term memory. 

\begin{equation}
\begin{split}
f_t &= \sigma(W_f \cdot [h_{t-1}, x_t]+b_f)\\
i_t &= \sigma(W_i \cdot [h_{t-1}, x_t]+b_i)\\
\hat{C}_t &= \text{tanh}(W_C \cdot [h_{t-1}, x_t]+b_C)\\
C_t  &= f_t \otimes C_{t-1} \oplus i_t \otimes  \hat{C}_t\\
o_t &= \sigma(W_i \cdot [h_{t-1}, x_t]+b_i)\\
h_t &= o_t\otimes\text{tanh}(C_t)\\
\end{split}\label{eq:LSTM}
\end{equation}

\subsection{
Autoencoder
}

The Encoder-Decoder-based models have been widely researched in different fields such as Natural Language Processing (NLP) and AD \cite{Sutskever2014SequenceNetworks, Zong2018DeepDetection}. In the unsupervised TSAD task, the \textit{seq2seq} model with an RNN-based Encoder-Decoder (Enc-Dec) has been suggested. The backbone of the \textit{seq2seq} is an encoder-decoder network, which can be RNNs, MLPs, or CNNs; the encoder is used to learn a new latent space with compressed feature representations, and a decoder network is used to decode the compressed features from this subspace to its original space \cite{Malhotra2016LSTM-basedDetection}. The compressed features from the encoder can be used not only to reconstruct the data but also for the AD algorithms as it contains the compressed information of the input data, which might be a long sequence or has very high dimensions \cite{Zong2018DeepDetection}.

In \cite{Liu2021DeepApproach}, a CNN-RNN-based framework was proposed by having a CNN module to extract spatial dependencies and an RNN module to capture temporal dependencies. In \cite{Vu2022LearningDetection}, a supervised AE was used to learn a latent space and several discriminative classifiers were employed to predict the anomalies in IoT devices. For time series data, \cite{Malhotra2016LSTM-basedDetection} proposed a framework based on an LSTM Enc-Dec and computed the anomaly score to measure the data deviation from the normal pattern. Similarly, \cite{Yoo2021RecurrentDetection} proposed a reconstruction-based LSTM Enc-Dec to model data with variable lengths, and the anomaly score was computed to detect the anomalies.

An adversarial-training-based Transformer was proposed in \cite{Tuli2022TranAD:Data} to enhance the performance of reconstruction-based anomaly detection for time series data. Additionally, a Transformer-based model that learned the Association Discrepancy to extract both local and global features in time series was introduced in \cite{Xu2022AnomalyDiscrepancy} for TSAD. Furthermore, \cite{Zhao2024SatelliteData} developed the Seasonal Decomposition Cross Attention Transformer for TSAD, which specifically addresses the issue of data non-stationarity by incorporating seasonal decomposition cross attention into its decoder to mitigate the data non-stationarity. While Transformer-based models have proven effective in various fields such as computer vision, natural language processing, and time series analysis \cite{Wen2023TransformersSurvey}, they typically require a large amount of training data. In contrast, RNNs, with their simpler architecture for temporal data modelling, can effectively capture temporal dependencies with less training data. Therefore, in this paper, RNNs are chosen as the backbone of the reconstruction module.

We advocate that all unsupervised Enc-Dec-based AD models focus on the reconstruction or prediction process, in which an anomaly score will be used to measure the quality, e.g., a higher score represents a poor reconstruction process and hence, is more likely to be an anomaly. 

The proposed RWPNN, in contrast, aiming at creating a probabilistic model for the latent space, is not restricted by the reconstruction process, which is more robust when limited training data with concept drift are present. Moreover, the high algorithm complexity of the existing WDE-based solutions caused by the high dimensionality of the original data can be neutralised, and the anomaly detection performance can be maximised in the proposed framework.

\subsection{Wavelet Density Estimation}
Density estimation can be classified into two categories  \cite{alma995345184401591}: parametric and nonparametric. The former, such as Gaussian Mixture Model, assumes that the data distribution is pre-defined. However, this assumption may not hold in many cases, especially when the data may have statistical properties shift in a non-stationary environment. Therefore, nonparametric methods, such as kernel density estimation, are preferred over parametric ones as they do not impose any restrictions on the form of the density.

The Orthogonal Series Estimator (OSE) is a nonparametric method that estimates densities efficiently, but it fails to capture the local features of the underlying density distribution. A better alternative, namely Wavelet Density Estimator (WDE), which inherits the computational advantage of OSE and also exploits the wavelet functions to enable local learning and manipulation of the estimated density, thus reflecting its local nature in time-frequency domains\cite{Vidakovic1999StatisticalStatistics}.

Given an unknown square-integrable density function $p(x)$, it can be estimated by a series of orthogonal basis functions in OSE such that $p(x)= \sum_{j\in\mathcal{J}} b_j \psi_j(x)$, with $b_j =\left\langle p, \psi_j \right\rangle$ being defined as the coefficient of the $j^{th}$ order of the basis function ${\psi_j},  \psi_j \in L^2(\mathbb{R})$, and $\mathcal{J} \in \mathbb{Z}$ is an appropriate set of indices. Since $p(x)$ is a density function, $b_{j}=\left\langle p, \psi_j \right\rangle = \int \psi_j (x)p(x)dx=E[\psi_j(X)]$. Given a group of independent random variable $X = \{ X_1, X_2, ..., X_N\}$ from this unknown distribution, an empirical counterpart of $b_{j}$ is given as $\hat{b}_{j}=\frac{1}{N} \sum_{i=1}^{N} \psi_j(X_i)$; the resulting approximated density $\hat p(x)$ can be calculated as: $\hat p (x)= \sum_{j} \hat b_j \psi_j(x)$.

The same concept also applies to WDE, such that the scaling function $\phi$ and wavelet function $\psi$ are used to replace the orthogonal basis functions: $\hat a_{j_0,k}=  \frac{1}{N} \sum_{i=0}^{N} \phi_{j_0,k}(X_i) $ and $\hat d_{j,k}= \frac{1}{N} \sum_{i=0}^{N} \psi_{j,k}(X_i) $, respectively; where $j_0$ indicates the coarsest scale or the lowest resolution of analysis, and $J$ refers to the finest scale or the highest resolution. Hence, the density can be approximated using one of the following three approaches: (i) using only scaling functions with $\hat p(x)= \sum_k \hat a_{j_0,k} \phi_{j_0,k}(x)$; (ii) using only wavelet functions with $\hat p(x)= \sum_{j=0}^{J} \sum_k \hat d_{j,k} \psi_{j,k}(x); $ and (iii) using both scaling and wavelet functions with $\hat p(x)= \sum_k \hat a_{j_0,k} \phi_{j_0,k}(x) + \sum_{j=j_0}^{J} \sum_k \hat d_{j,k} \psi_{j,k}(x)$.

\subsection{Wavelet Neural Networks}
The Wavelet Neural Network (WNN) in \cite{zhang1992wavelet}, combines wavelets and neural networks, which enables them to exploit the key advantages of wavelet theory, such as multiresolution analysis, effective localisation in time-frequency domains, and sparse representation of functions. It has been applied to various tasks, such as function learning and time series prediction \cite{Walter1995WaveletLearning, Chen2006Time-seriesNetwork}.

Most of the existing studies on NNs do not model the probability density function of the data using wavelets, but rather use wavelets in a non-probabilistic way. For instance, wavelets are commonly used in signal denoising and multi-resolution decomposition \cite{Liu2022T-Wavenet:Analysis, Chen2022AForecasting}. There are few studies that integrate wavelets into the probabilistic framework using WDEs, as conventional WDEs are computationally costly due to the absence of analytic closed-form solutions.

\section{Wavelet Probabilistic Neural Networks}\label{sec:wpnn}
Probabilistic neural networks (PNNs) are a type of neural network that uses Gaussian basis functions to estimate the class-conditional densities of the input data \cite{Specht1988ProbabilisticMemory}. The density estimation for a class $c \in C$ is given by: $\hat p_c(\mathbf{x}| c)=\frac{1}{(2\pi)^{n/2}  \sigma^n}\frac{1}{N_c} \sum_{i=1}^{N_c}\exp ( -\frac{1}{2\sigma^2}(\mathbf{x}-\mathbf{x}_{c,i})^T(\mathbf{x}- \mathbf{x}_{c,i}) )$ where $\mathbf{x}_{c,i}$ is the $i$-th training data in the class $c$, $\mathbf{x}$ is the test set, $n$ is the data dimension and $N_c$ is the length of training set in $c$. The smoothness of the density estimation is controlled by $\sigma$.

Data stream classification is a challenging task that involves dealing with unconstrained data length and possible concept drift over time. PNNs suffer from performance degradation and computational inefficiency in this setting, as they depend on the size of the training data \cite{Garcia-Trevino2024WaveletNetworks}. A novel WDE-based Probabilistic Neural Network (WPNN) we proposed in \cite{Garcia-Trevino2024WaveletNetworks} overcomes these limitations. This network only needs to store the last data point and has a constant computational and space complexity regardless of the window size.

\subsection{Radial B-Spline Scaling Function}\label{sec:b_spline}
Unlike conventional WDE suffering from high computational burden due to no analytic closed-form solutions, WPNN utilises a novel multidimensional scaling function $\Phi$, namely \textit{Radial $B$-spline scaling functions}, which is defined as:

\begin{equation} 
\Phi_{j_0,k}(\mathbf{x})=2^{\frac{nj_0}{2}}N_m \big( \lVert (2^{j_{0}}\mathbf{x}-\mathbf{k}) \rVert + \frac{m}{2}  \big)
\label{eq:bspline}\end{equation} 

\noindent
where $n \in \mathbb{Z}$ is the dimension of $\mathbf{x}$ , $j_0 \in \mathbb{Z}$ is the dilation parameter, and $\mathbf{k} \in \mathbb{Z}^n$ is a vector of translation parameters.

The $m$-th order $B$-spline $N_m(x)$ can be obtained by using convolution, where $N_1(x)$ is $1$ when $\text{for }0\leq x<1$ and $0$ otherwise:

\begin{equation} 
N_m(x)=\int_{-\infty}^{\infty} N_{m-1}N_1(t)dt= \int_{0}^{1}N_{m-1}(x-t)dt
\label{eq:bspline_Nm}\end{equation} 

The resulting analytic closed-form solution of the $m$-th order $B$-spline $N_m(x)$ is given in Table \ref{tbl:closed_form}, where $\phi(x)$ is the different orders of $N_m(x)$. 

\begin{table}[!ht]
\centering
\caption{Closed-form solution for $B$-spline functions.}
\label{tbl:closed_form}
\scalebox{0.9}
{
\begin{tabular}{cll}
\hline
\textbf{Order}& \textbf{$B$-spline function} \\
\hline
\begin{tabular}[c]{@{}c@{}}Linear\\ m=2\end{tabular}
& 
\(
\phi(x)=\begin{cases}x & \hphantom{1111111111111111111}\text{for }0\leq x<1 \\
2-x & \hphantom{1111111111111111111}\text{for }1\leq x<2 \\
0 & \hphantom{1111111111111111111}\text{otherwise } \\
\end{cases}\nonumber
\)
\\
\hline
\begin{tabular}[c]{@{}c@{}}Quadratic\\ m=3\end{tabular}
& 
\(
\phi(x)=\begin{cases}\frac{1}{2}x^2 & \hphantom{111111111111}\text{for }0\leq x<1 \\
\frac{3}{4}-(x-\frac{3}{2})^2 & \hphantom{111111111111}\text{for }1\leq x<2 \\
\frac{1}{2}(x-3)^2 & \hphantom{111111111111}\text{for }2\leq x<3 \\
0 & \hphantom{111111111111}\text{otherwise } \\
\end{cases}\nonumber
\)
\\
\hline
\begin{tabular}[c]{@{}c@{}}Cubic\\ m=4\end{tabular}
& 
\(
\phi(x)=\begin{cases}
\frac{1}{6}x^{3} & \text{for }0\leq x<1 \\
\frac{1}{6}(-3x^{3}+12x^2-12x+4) & \text{for }1\leq x<2 \\
\frac{1}{6}(3x^{3}-24x^2+60x-44) & \text{for }2\leq x<3 \\
\frac{1}{6}(4-x)^{3} & \text{for }3\leq x<4 \\
0 & \text{otherwise } \\
\end{cases}\nonumber
\)
\\
\hline
\end{tabular}
}
\end{table}

\subsection{Density Estimation}
A density function $p(\mathbf{x})$ in $L^2(\mathbb{R})$ can be approximated using a basis of scaling functions. The coefficient $w_{j_0,k}$, which is defined as the inner product of $p(x)$ and the frame function $\{\Phi_{j_0,k}\}_{j_0,k \in \mathbb{Z}}$, is given as: 

\begin{equation} 
w_{j_0,k}=\left\langle p, \Phi_{j_0,k} \right\rangle = \int p(\mathbf{x})\Phi_{j_0,k}(\mathbf{x})d\mathbf{x}\label{eq:w_coef}
\end{equation}

Equation \ref{eq:w_coef} is further expressed as $w_{j_0,k} = \mathbb{E}[\Phi_{j_0,k}(\mathbf{x})]$ due to $p(\mathbf{x})$ is a density function. Given a group of independent random variables $X_i$, $i \in [1,N]$ from the unknown distribution $p$, the empirical counterpart of $w_{j_0,k}$ is given as:

\begin{equation} 
\hat w_{j_0,k}=\frac{1}{N} \sum_{i=0}^{N} \Phi_{j_0,k}(X_i)  
\label{eq:coef_update}\end{equation}

The estimated density $\hat p(\mathbf{x})$ at resolution $2^{j_0}$ is shown in Equation \ref{eq:pdf}, where the $\mathbf{x}$ is normalised within the interval $[0,1]^n$, $n$ denotes the corresponding feature dimension:

\begin{equation} 
\hat p(\mathbf{x}) = \sum_{k} \hat w_{j_{0},k} \Phi_{j_{0},k}(\mathbf{x})   
\label{eq:pdf}
\end{equation}

\subsection{Coefficient Update}\label{sec:wpnn_coef_update}
Following the discussion in \cite{GARCIATREVINO2019111}, Equation \ref{eq:coef_update} can be optimised in a recursive way, such that each coefficient is updated by processing one data point at a time, given $j_0$ and $k$:

\begin{equation} 
\begin{aligned}
\hat w^{t} &= (1-\alpha)\hat w^{t-1} + \alpha\big(\Phi_{j_0,k}(\mathbf{x}_t)  \big)
\\
&=(1-\alpha)\hat w^{t-1} + \alpha\Big( 2^{\frac{nj_0}{2}}\phi\big( \lVert (2^{j_{0}}\mathbf{x}_t-\mathbf{k}) \rVert + \frac{m}{2} \big)  \Big)
\end{aligned}
\label{eq:online_nonstat}
\end{equation}

Note that Equation \ref{eq:online_nonstat} is designed and used for the non-stationary environment, where $\alpha$ is the forgetting factor, which relates to a sliding window with size $P$ to analyse the latest $P$ data points. The support of the $B$-spline scaling function $\Phi(\mathbf{x})$ with any dilation and translation parameters $j_0$ and $k$ is $[2^{-j_0}(k-\frac{m}{2}), 2^{-j_0}(k+\frac{m}{2})]$. Note that the number of frame functions required to form a frame within the support is $2^{j_0} + 2u +1$, where $u=1$ for \textit{Linear} and \textit{Quadratic} order $B$-spline and $u=2$ for \textit{Cubic} order. The resulting multivariate translation vector $\mathbf{k}$ is all the combinations of the translation parameter $k$ across the data dimension $n$, where $k$ is then given as $k \in \{-u, ..., 0, 1, 2, ..., 2^{j_0}+u\}$. 

\subsection{Density Estimation and Classification}\label{sec:wpnn_density_estimation}
Given that the network coefficients $\hat{w}_{j_0,k}$ ($\hat{w}$ for simplification here) with the specific values of $j_0$ and $k$ are trained by using Equation \ref{eq:pdf}, there exists a set of $W$ such that $W = [w_1, w_2, ..., w_c], c \in \mathbb{Z}$. For the test data $\mathbf{x}$, the class conditional probability density function (PDF) $\hat{p}(\mathbf{x}|\mathcal{C}_{c})$ for the class $c$ can then be computed using \ref{eq:pdf}:


\begin{equation} 
\begin{aligned}
\hat p_{\mathcal{C}_{c}}(\mathbf{x}|\mathcal{C}_{c}) &= \sum_{k}\hat{w_c} \cdot \Phi_{j_0}(\mathbf{x})\\
&= \sum_{k}{\hat{w_c} \cdot 2^{\frac{nj_0}{2}}}  \phi(\left\Vert 2^{j_0}\mathbf{x}-k \right\Vert+\frac{m}{2})
\end{aligned}
\label{eq:class_prob}
\end{equation}

Therefore, for $C$-class of data, the decision rule is defined as $d(\mathbf{x}) = \arg\max_{c} \{ P(\mathcal{C}_{c}) \hat p_{\mathcal{C}_{c}}(\mathbf{x}| \mathcal{C}_{c})\}, \quad c=1,\ldots,C$, where $P(\mathcal{C}_{c})$ is the prior and it is assumed to be uniform.

\subsection{Hyper Parameters}
WPNN has three main hyperparameters that need to be optimised. The first one is the order of the $B$-spline scaling functions, which determines the smoothness of the density estimation process. The second one is the resolution parameter $j_0$, which also affects the smoothness. It should be selected based on the complexity of the density function and the size of the data set. A larger $j_0$ may reveal more details but also increase the risk of overfitting. The third parameter, $\alpha$, is designed to update the network coefficients for non-stationary data and it corresponds to the inverse of the sliding window size $P$, i.e., $\alpha= {1/P}$. Note that the selection of $\alpha$ value can be perceived as choosing an appropriate value that reflects the rate of data variation; if the data statistical properties change rapidly, then a higher $\alpha$ value is expected, where data located closer to the current timestamp will be focused.


\subsection{Limitations}
Despite the outstanding performance of WPNN in terms of time complexity and classification results for the non-stationary environment, it still has some limitations: (i) WPNN requires all the classes to be known for training (see Section \ref{sec:wpnn_density_estimation}); however, in TSAD tasks, estimating the PDF for both classes and applying the decision rule might be redundant. (ii) the hyperparameter tuning process is sub-optimal, for example, in a TSAD task, we need to train WPNN $2*|\mathcal{M}_m|*|J|*|\Gamma|$ times, where $2$ is the number of classes, $|\mathcal{M}_m|,|J|,|\Gamma|$ are the cardinality of the possible hyperparameter sets for the order of $B$-spline $m$, resolution $j_0$, and forgetting factor $\alpha$, respectively. (iii) WPNN directly models the PDF of the original feature space of the input data $\mathbf{x}$, which becomes infeasible when hundreds of features are involved, as it significantly increases the computational complexity.

To address these limitations while preserving the benefits of WPNN in detecting anomalies and estimating the density function with low and constant complexity in the non-stationary environment, we propose an unsupervised TSAD method that consists of two modules: (i) \textit{SREnc-Dec} that learns a latent space from the input data, and (ii) an advanced probabilistic network (\textit{MRWPN}) that models the PDFs of the latent space and provides multiple views to the data variation. Note that these two steps remove the data dimensionality restriction of WPNN as data with a very long sequence and high dimensions is represented by temporal features in a lower dimensional space with a shorter length; and the step of tuning the $|\Gamma|$ set of WPNNs is avoided as all the possible views are generated in an ensemble view at the same time without formulating extra models.

Table \ref{tbl:complexity_analysis} shows the time complexity comparison of processing a sequence with length $L_o$, where $L_{\{o,t\}}$ are the lengths of the original and the compressed temporal sequences, respectively; $n_{\{o,t\}}$ are the data dimensions; $n_c$ is the number of classes; $M_c$ is the cardinality of the relevant frame function, whose evaluation of Equation \ref{eq:bspline} are not zero; and $m$ is the order of $B$-spline. In \textit{WPNN}, the original sequence with length $L_o$ and $n_o$ features is directly fed to the framework, and the forgetting factor $\alpha$ is fixed. This becomes inefficient when $n_c$ is very large. In the proposed RWPNN, the algorithm complexity can be reduced by using the proposed two modules: (i) \textit{SREnc-Dec}, which extracts meaningful compressed temporal features with shorter length  $L_t$ , $(L_t << L_o)$ and lower feature dimensions $n_t$, and (ii) \textit{MRWPN}, which is an ensemble-based network that estimates the PDFs using the latent features from \textit{SREnc-Dec} with $|\Gamma|$ different views to model different rates of data variation.

\begin{table}[!htp]
\centering
\caption{Time Complexity Comparison between WPNN \cite{Garcia-Trevino2024WaveletNetworks} and the proposed RWPNN.}
\label{tbl:complexity_analysis}
\begin{tabular}{ccc}
\multicolumn{3}{c}{Time Complexity}                                \\ \hline \hline
\textit{Step}    & \textit{WPNN}             & \textit{MRWPN}               \\ \hline
Coefficient update & O($L_o M_c n_o m$)       & O($L_t M_c n_t m$) \\ 
Density Estimation       & O($L_o(n_c + M_c n_o m)$) & O($L_t M_c n_t m$) \\\hline \hline
\end{tabular}
\end{table}

\section{Methodology}\label{sec:method}

The proposed RWPNN consists of two modules: (i) an RNN-based encoder-decoder that learns a new latent space and reconstructs the latent features back to its original data space, and (ii) a novel ensemble multi-receptive-field wavelet probabilistic network (\textit{MRWPN}) that models the probability densities of the latent space with different perspectives to different rates of data variation by using all the possible $\Gamma = [\alpha_1,\alpha_2,...,\alpha_\gamma], \Gamma \in \mathbb{Z}$ at the same time (see Fig. \ref{fig:rwpnn_flowchart}). We note that utilising all the possible forgetting factors at the same time provides $|\Gamma|$ perspectives associated with different rates of variation, and therefore, can be perceived as having multiple receptive fields. The encoder and decoder form the time series reconstruction module (\textit{SREnc-Dec}), while the \textit{MRWPN} module performs the anomaly detection task.

\subsection{Stacked Recurrent Encoder-Decoder Module}
\begin{figure*}[htp]
\centering
\includegraphics[trim = 0mm 0mm 0mm 0mm,clip,width=0.85\linewidth]{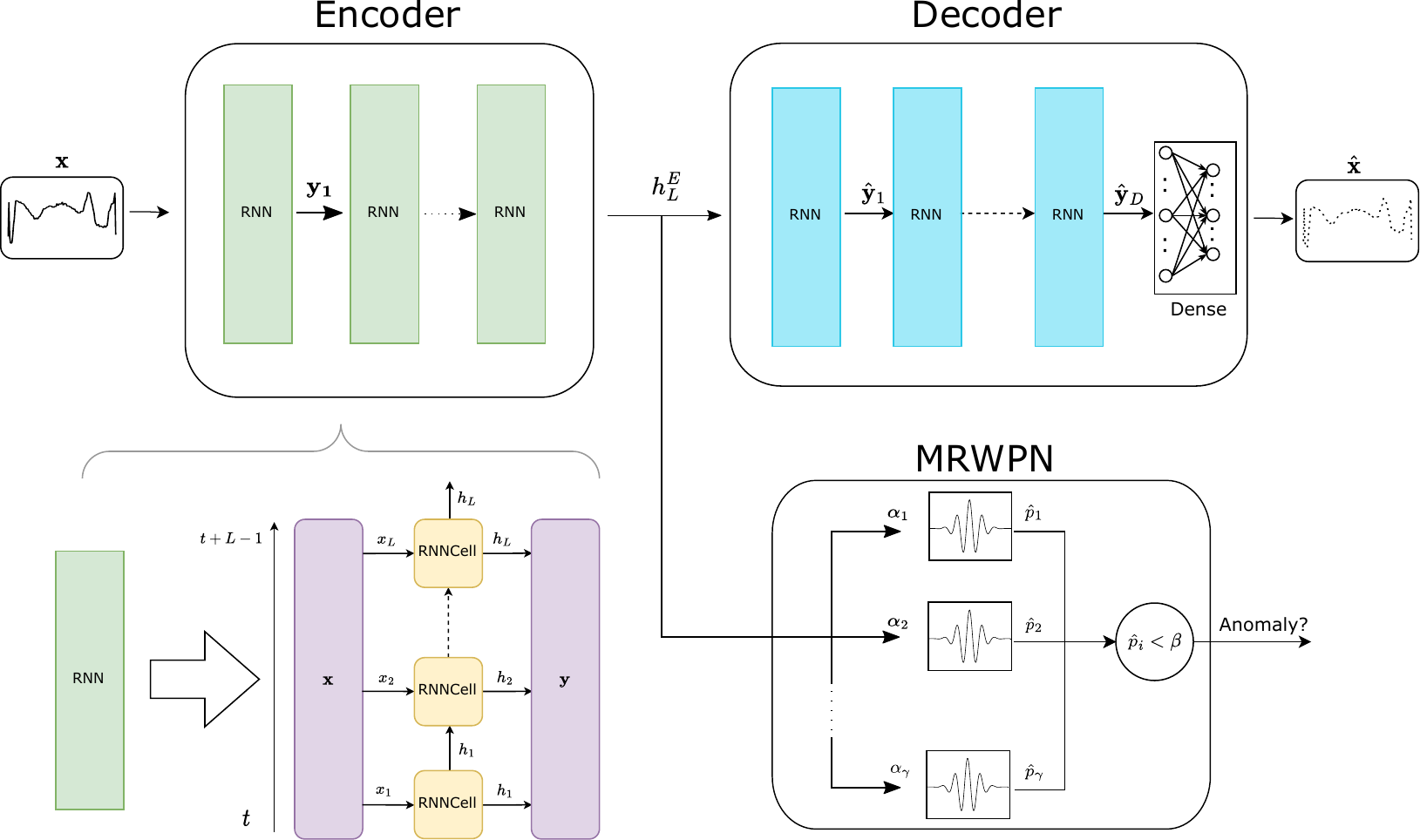}
\caption{The flowchart of the proposed RWPNN. The temporal feature extractor block consists of an RNN-based Encoder and a Decoder to compress and reconstruct the input signal. And a Multi-receptive-field Wavelet Probabilistic Network (MRWPN) creates an ensemble probabilistic model that handles different rates of data variation in the non-stationary environment for anomaly detection.}\label{fig:rwpnn_flowchart}
\end{figure*}

The reconstruction module is a stacked Recurrent Encoder-Decoder (\textit{SREnc-Dec}) network utilising RNN with variable output feature dimensions. $L$-RNNCells are connected to form one RNN shown in Fig. \ref{fig:rwpnn_flowchart}. The output $y_e$ from the $e$-th RNN ($e = [1,2,...,E], E\in\mathbb{Z}$) is fed to the $(e+1)$-th RNN. We note that this configuration can be perceived as the stacked RNNs when configuring the network, and the output feature dimension of each RNN can be adjusted, which increases the model's ability to learn the complex hidden representation of the time series data. 

The \textit{Encoder}, which consists of $E$ stacked RNNs, will pass the hidden state $h^E_L$ from the final layer of the RNN as the input to the \textit{Decoder} and \textit{MRWPN}; the former one who has $D\in\mathbb{Z}$ stacked layers, will map $h^E_L$ to the input space of RNN and learn to reconstruct from $h^E_L$ to $\mathbf{x}$ using the stacked-RNN and a dense layer; and the latter one can detect the anomalies by estimating the class conditional probability densities of the current data: $\hat{p}(h^E_L|\mathcal{C}_{\text{normal}})$ in terms of multiple PDFs with $|\Gamma|$ different views being computed. The objective function for RWPNN is set to minimise the error between $\mathbf{x}$ and $\mathbf{\hat{x}}$ using the mean absolute error (MAE). Note that the term RNN can be any architecture of the recurrent neural networks, e.g., Vanilla RNN, LSTM, or GRU; in this paper, LSTM is used for demonstration.

\subsection{The Proposed MRWPN Module}

\begin{figure}[htp]
\centering
\includegraphics[width=0.9\linewidth]{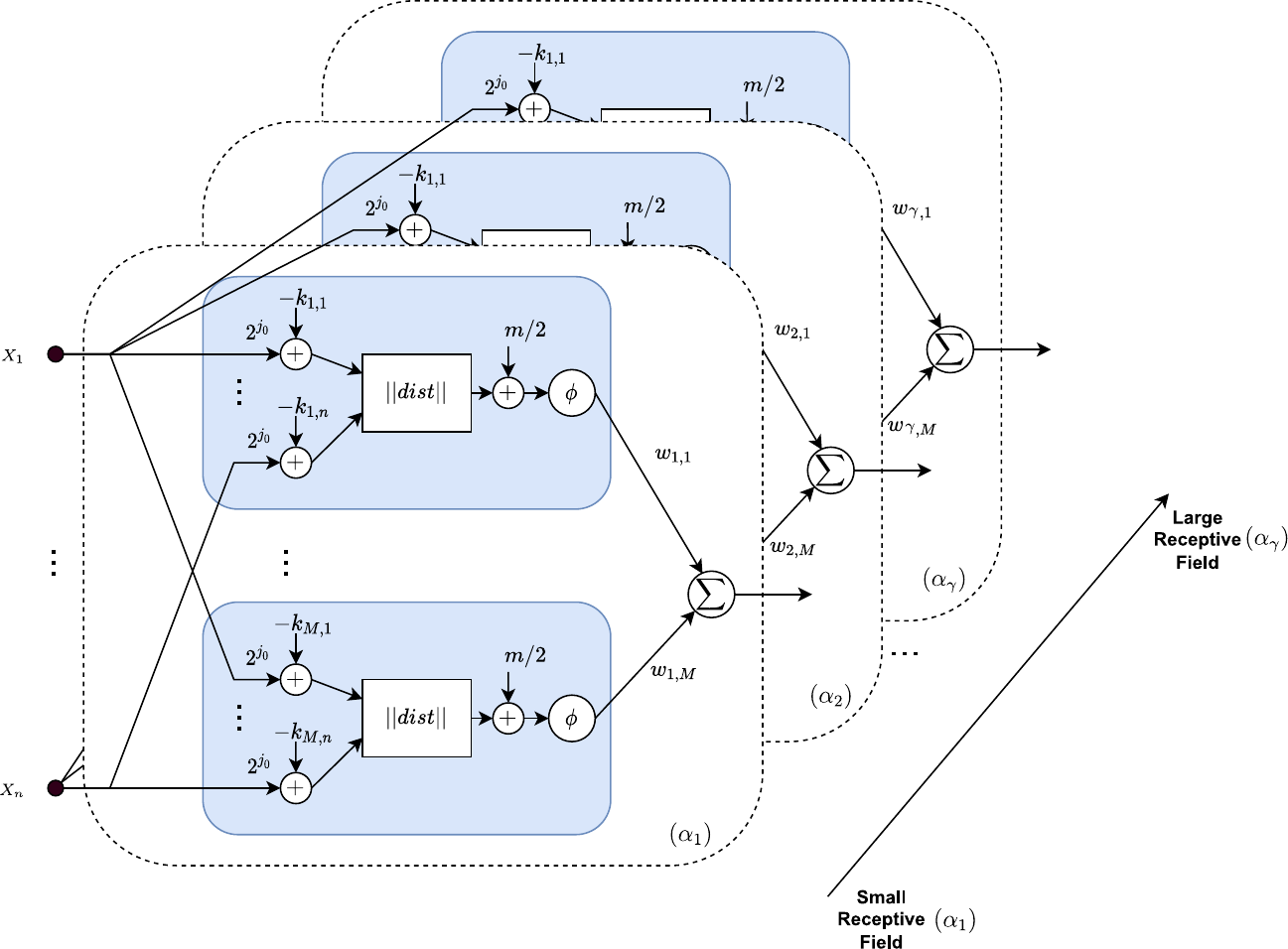}
\caption{
Visualisation of the \textit{MRWPN} module, which provides $|\Gamma|$ views at the same time. Each dashed module refers to an MRWPN with a specific value of $\alpha$, and different values of $\alpha$ are used to construct the ensemble view. 
} \label{fig:rwpnn_multi_recepticve_wpnn}
\end{figure}

The structure of the \textit{MRWPN} module is depicted in Fig. \ref{fig:rwpnn_multi_recepticve_wpnn}. It computes the PDF across $|\Gamma|$ different scales of $\alpha$, where $\Gamma = [\alpha_1, \alpha_2, ..., \alpha_\gamma]$ and $|\cdot|$ denotes the cardinality of $\Gamma$. Each module (in dashed line) is characterised by a specific value of $\alpha$, and each blue block inside evaluates Equation \eqref{eq:class_prob} with a specific value of $k$. By considering all the possible views characterised by different values of $\alpha$, an ensemble view is formed.

The proposed RWPNN has the following advantages: (i) It is trained in an unsupervised way using the normal class of data, i.e., $\mathcal{C}_{\text{normal}}$. (ii) \textit{SREnc-Dec} can learn a compressed temporal feature space which can further reduce the time complexity of \textit{MRWPN} module as the original input sequence is characterised by a smaller temporal pattern. Therefore, the curse of dimensionality in WPNN and WDEs has been neutralised. (iii) Multiple $\alpha_{[1,2,...\gamma]}$ are used to provide an ensemble view, targeting the different rates of data variation. Hence, the hyperparameter tuning step for $\alpha$ from WPNN can be avoided.

Note that existing methods, e.g., RNN-based AE and VAE, rely on modelling the input and latent space via sequence reconstruction, prediction, or Gaussian-based models, which can be easily affected in the non-stationary environment; \textit{MRWPN}, on the other hand, models the latent space in a different probabilistic way using \textit{MRWPN}, which is a more robust solution for density estimation in the non-stationary environment.

\subsection{Wavelet Optimisation}

\subsubsection{Initialisation}
As discussed in Section \ref{sec:wpnn_coef_update}, four parameters are needed for \textit{MRWPN}: (i) $m$, the order of $\Phi(x)$ from Equation \ref{eq:bspline}. (ii-iii) $k$ and $\mathbf{k}$, the translation parameter and the vectorised version for the multidimensional frame $\Phi(x)$. (iv) $\hat{w}$, the network parameter such that $\hat{w}\in \mathbb{R}^{|k|,|\Gamma|}$, where $|\cdot|$ is the cardinality of $k$ and $\Gamma$.

\subsubsection{Parameter Optimisation}
The network parameters $\hat{w}$ of the \textit{MRWPN} module can be updated using Equation \ref{eq:MRWDE_nonstat} and Algorithm \ref{algo:MRWDE_train} without using back-propagation:

\begin{equation} 
\begin{aligned}
\hat w &=(1-\Gamma)\hat w + \Gamma\Big( 2^{\frac{nj_0}{2}}\phi\big( \lVert (2^{j_{0}} \cdot h^E_L -\mathbf{k}) \rVert + \frac{m}{2} \big)  \Big)
\end{aligned}
\label{eq:MRWDE_nonstat}
\end{equation}

where $\Gamma$ is the set containing different scales of the receptive field $\alpha$, 
$n$ is the feature size of $h^E_L$, 
$j_0$ is the resolution parameter, $\mathbf{k}$ is the vector of the translation parameters, $m$ is the order of the $B$-spline scaling function, and $\setminus$ in \textit{$\mathbf{f} \setminus \mathbf{b}$} removes the elements in $\mathbf{b}$ from $\mathbf{f}$. 

Note that there are in total $(2^{j_0} + 2u + 1)^n$ frame functions for an $n$-dimensional scaling function $\phi(x)$ (see Section \ref{sec:wpnn} and \cite{GARCIATREVINO2019111} for detailed discussions), where the support for each dimension $d$ is given as $[2^{-j_0}(k_d - \frac{m}{2}), 2^{-j_0}(k_d + \frac{m}{2})]$. To further optimise the computational complexity, Algorithm \ref{algo:algorithm_find_relevant_frame} is employed to identify the indices $b$ of the relevant frame functions where the evaluation of $\phi(x)$ is not zero.


\begin{algorithm}
\scriptsize
\caption{\footnotesize \sffamily MRWPN on-line non stationary training ($\mathbf{x}, \hat{w_c},j_0,m,M, \mathbf{k},\Gamma$) }
\label{algo:MRWDE_train}
\DontPrintSemicolon
\KwIn{$\mathbf{x}$: Training data $\mathbf{x}$ from the Encoder; $\hat{w}$: The set of coefficients; $j_0$: The resolution parameter; $m$: The order of the $B$-spline function employed as scaling function; $M$: The number of frame functions in [0,1]; $\mathbf{k}$: Translation vectors for the frame functions; $\Gamma$: The vector of the forgetting factors.}
\KwOut{$\hat{w}$: Set of updated network coefficients.}

$\mathbf{b}$ = find relevant frame($\mathbf{x}$,$j_0$, $\mathbf{k}$,$M$)\\
$M_c = |\mathbf{b}|$\\
\For{$l\leftarrow 1$ \KwTo $M_c$ }{
$\hat w_{b_l}=(1-\Gamma)\hat w_{b_l}+  \Gamma \Big(2^{\frac{nj_0}{2}}\phi\big( \lVert (2^{j_{0}}\mathbf{x}_{t}-\mathbf{k}_{b_l}) \rVert + \frac{m}{2} \big) \Big)$ \\ 

}
$\mathbf{f} = [1,2,...,M]$;\\
$\mathbf{f}\char`\\\mathbf{b} $ = $[1,2,...,M]\char`\\[b_1, b_2,...,b_{M_c}]$\\
\For{$l\leftarrow 1$ \KwTo $M - |b|$ }{
$\hat w_{f_l} = \hat w_{c,f_l} - \Gamma \cdot {\hat w_{c,f_l}}$
}
\end{algorithm}
\normalsize

\begin{algorithm}
\scriptsize
\caption{\footnotesize \sffamily find relevant frame($\mathbf{x}$,$j_0$,$m$,$\mathbf{k}$,$M$) }
\label{algo:algorithm_find_relevant_frame}
\DontPrintSemicolon
\KwIn{$\mathbf{x }$: Training data $\mathbf{x}$ from the Encoder; $j_0$: The resolution parameter; $m$: The order of the $B$-spline function employed as scaling function; $\mathbf{k}$: Translation vectors for the frame functions in [0,1]; $M$: Number of frame functions in [0,1].}
\KwOut{$\mathbf{b}=\{b_1, b_2, ..., b_B \} $: Set of indices of the translation vectors for the frame functions relevant for $\mathbf{x} $.}
$n = dim(\mathbf{x })$\\
$\mathbf{h}=[h_1,h_2,...,h_n]=[0,0,...,0]$\\
$\mathbf{x}=\mathbf{x }$\\
\For{$j\leftarrow 1$ \KwTo $M$ }{
    \For{$d\leftarrow 1$ \KwTo $n$ }{
        \If{$\mathbf{x_d} \geq 2^{-j_0}(\mathbf{k}_j-\frac{m}{2})$ and $\mathbf{x_d} \leq 2^{-j_0}(\mathbf{k}_j+\frac{m}{2})$}{
        $h_d \leftarrow h_d +1$\\ 
        \eIf{d == 1}{
            $A_{d,h_d} \leftarrow j$\\
            }{
            $A_{d,h_d} \leftarrow (j-1)M^{d-1}$\\
            }
        }
    }
}
$\mathbf{b} = A_{1,*}$\\
\For{$d \leftarrow 1$ \KwTo $n-1$ }{
    $\mathbf{g} \leftarrow \{\}$\\
    \For{$i \leftarrow 1$ \KwTo length($\mathbf{b}$) }{
        \For{$p \leftarrow 1$ \KwTo columns($A$)}{
            $\mathbf{g} \leftarrow concatenate(\mathbf{g}, (b_i + A_{d+1,p}))$\\
        }
    }
    $\mathbf{b} \leftarrow \mathbf{g}$\\ 
}

\end{algorithm}
\normalsize

\subsection{Unsupervised Anomaly Detection}
RWPNN is trained in an unsupervised manner, utilising the normal class of the data. Note that for simplicity, from here we write $\hat{p}(\mathbf{x})$ and $h^E$ instead of $\hat{p}(\mathbf{x}|\mathcal{C}_{\text{normal}})$ and $h^E_{L}$, respectively, unless otherwise specified, where $\hat{p}(\mathbf{x}) = \hat{p}_{\{ 1,2,...,\gamma\}}(\mathbf{x})$ 
to denote the collection of the ensemble views. 
The coefficients of \textit{MRWPN} is updated using $h^E_{\text{train}}$ during the training stage; for test set, \textit{MRWPN} estimates $\hat{p}(\mathbf{x}_{\text{test}})$ using $h^E_{\text{test}}$ providing $\Gamma$ different views of PDFs, see Algorithm \ref{algo:get_prob} for details.



\begin{figure}[htp]
\centering
\includegraphics[width=0.9\linewidth]{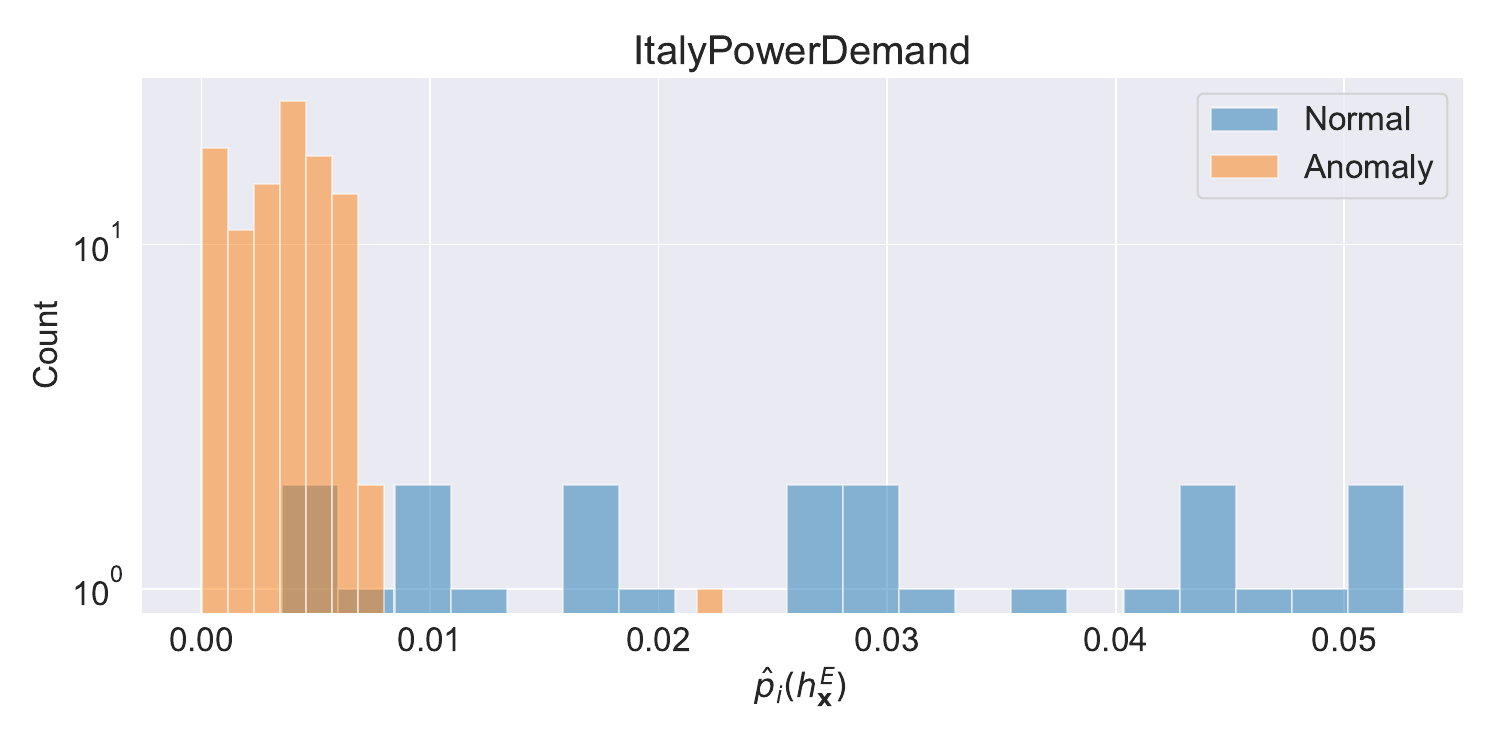}
\caption{Distribution of the estimated $\hat{p}_i(h^E_{\mathbf{x}})$ in Dataset \textit{ItalyPowerDemand}; the normal and anomaly data can then be separated by selecting an appropriate threshold $\beta$.} \label{fig:pdf_count}
\end{figure}

Following the observation in \cite{Garcia-Trevino2024WaveletNetworks}, the class decision can be made to classify $\mathbf{x}$ as an anomaly if $\hat{p}(h^E_{\mathbf{x}})$ falls below a threshold. Fig. \ref{fig:pdf_count} depicts a histogram of the estimated densities for the normal and anomaly data from dataset \textit{ItalyPowerDemand} in \cite{Anh2018TheArchive}. It is clear to see that the anomalies tend to lie at the region with the lower value of $\hat{p}(h^E_{\mathbf{x}})$; therefore, a joint optimisation can be made to find the best $i$-th \textit{MRWPN} model that describes the latent space while maximising the F1-score by selecting an appropriate threshold $\beta$: 

\begin{equation}
\operatorname*{argmax}_{i \in \{1,..,\gamma\}, \beta} \text{F1}\{\hat{p}_i(h^E_{\mathbf{x}}) < \beta\}
\label{eq:decision_alpha}
\end{equation}

where
$i$ indicates the index of each ensemble view, each associated with unique values of $\alpha$.
The class decision for computing the F1-score is based on whether $\hat{p}_i(h^E_{\mathbf{x}_\text{test}})$ falls below a threshold $\beta$. Here, $h^E_{\mathbf{x}_\text{test}}$ is the encoded representation of the test sequence $\mathbf{x}_{\text{test}}$, computed by the Encoder of the \text{SREnc-Dec}. The test sequence $\mathbf{x}$ contains an anomaly if $\hat{p}_i(h^E_{\mathbf{x}_\text{test}}) < \beta$. 
Note that $\beta$ and $i$ can be determined using a validation set\footnote{
There are various methods to determine the optimal values, such as those discussed in \cite{An2015VariationalProbability, Malhotra2016LSTM-basedDetection, Guo2018MultidimensionalApproach}. We adopted the approaches discussed in \cite{Malhotra2016LSTM-basedDetection} and \cite{An2015VariationalProbability}.
}.

We further note that, by analysing the latent features $y^E$ from the last layer of the Encoder, a statistical pattern $\hat{p}(y^E)$ can be formed, which captures the variation of $\mathbf{x}$ in terms of PDFs over time. This allows for the analysis of subtle deviations from normal patterns, which can be useful for implementing an early warning system.


\begin{algorithm}
\scriptsize
\caption{\footnotesize \sffamily MRWPN Evaluation ($\mathbf{x},\mathbf{K},\mathbf{w},  j_0,m,\mathbf{k},M$) }
\label{algo:get_prob}
\DontPrintSemicolon
\KwIn{$\mathbf{x}$: Test set from the Encoder; $\hat{w}$: Trained network coefficients; $j_{0}$: The resolution parameter;  $m$: The order of the $B$-spline function; $\mathbf{k}$: Translation vectors for the frame functions in [0,1]; $M$: number of frame functions in [0,1].}
\KwOut{$\mathbf{\hat p}$: Set of probability densities for the testing data $\mathbf{x}$.}
\For{$i\leftarrow 1$ \KwTo $N_{\text{\upshape testing}}$ }{
$\mathbf{b}$ = find relevant frame($\mathbf{x}_{i}$,$j_0$, $\mathbf{k}$,$M$)\\
$M_c = |\mathbf{b}|$\\
\For{$l\leftarrow 1$ \KwTo $M_c$ }{
$z_{l}=2^{\frac{nj_0}{2}} w_{b_l} \phi(\left\Vert 2^{j_0}\mathbf{x}-\mathbf{k}_{b_l} \right\Vert+\frac{m}{2})$ \\ 
}
$\hat p(\mathbf{x}_i| \mathcal{C}_c )=\sum_{l=1}^{M} z_{b_l}  $\\
}
\end{algorithm}
\normalsize

\section{Experiments}\label{sec:result}

RWPNN is assessed using $45$ real-world time series datasets from the UCR and UEA time series data repositories \cite{Anh2018TheArchive, Bagnall2018The2018}, which include univariate and multivariate TS data collected from different applications. We mainly selected those datasets with binary classes and used one class as the training set. Moreover, we further select some healthcare-related datasets from these two repositories, train on the normal class, and detect the pathological class. 

Furthermore, we select a five-week-long server machine dataset (SMD) from \cite{Su2019RobustNetwork}, which comprises $28$ entities, $n=38$ features, $1e6$ data points in total and $4.2\%$ of anomalies. As discussed in \cite{Su2019RobustNetwork}, the data is preprocessed to have the sequence length $L = 100$.

The normal class of data are divided into $(1-\mathcal{P})$ of the training set $\mathcal{D}_{\text{train}}$, where $\mathcal{P} \in [0,1]$ denotes the proportion of dataset; and the remaining $\mathcal{P}$ of the normal data are further divided into $(1-\mathcal{P})$ of the validation set $\mathcal{D}_{\text{v}_1}$ for early stopping criteria, and the rest of the data forms part of the test set $\mathcal{D}_{\text{test}}$. A similar approach is applied to the anomaly class, where $(1-\mathcal{P})$ of data forms another validation set $\mathcal{D}_{\text{v}_2}$ and the remaining part forms the test set $\mathcal{D}_{\text{test}}$. $\mathcal{D}_{\text{v}_1 + \text{v}_2}$ are used for finding an optimal threshold in TSAD. 
Table \ref{tbl:data_summary} shows examples of dataset when $\mathcal{P}= 0.2$. Note that data are normalised into $[0,1]^n$. 

\begin{table}[!htp]
\centering
\caption{Dataset summary when the test percentage $\mathcal{P} = 0.2$, which includes the total size of the data, sequence length ($L$), number of features ($\mathbf{n}$), size of $\mathcal{D}_{\text{train}}$, $\mathcal{D}_{\text{v}_1 + \text{v}_2}$, $\mathcal{D}_{\text{test}}$, and the anomaly percentage $\mathcal{A}_{\text{perc}}$.}
\label{tbl:data_summary}
\scalebox{0.8}
{
\begin{tabular}{ccccccccl}
\hline \hline
\textbf{Dataset}       & \textbf{Data Size} & {$\mathbf{L}$} & $\mathbf{n}$ & $\mathbf{|\mathcal{D}_{\text{train}}|}$ & $\mathbf{|\mathcal{D}_{\text{v}_1 + \text{v}_2}|}$ & $\mathbf{|\mathcal{D}_{\text{test}}|}$ & $\mathcal{A}_{\text{perc}}$ \\ \hline
\textit{FaceDetection} & 9414               & 62            & 144           & 3765                           & 4518                                      & 1131                          & 0.50          \\
\textit{Heartbeat}     & 409                & 405           & 61            & 236                            & 138                                       & 35                            & 0.28          \\
\textit{ProxPhalOC}    & 891                & 80            & 1             & 484                            & 324                                       & 83                            & 0.32          \\
\textit{Wafer}         & 7164               & 152           & 1             & 5121                           & 1633                                      & 410                           & 0.11          \\ \hline
\end{tabular}
}
\end{table}

The performance of the proposed RWPNN is evaluated against several unsupervised benchmark algorithms covering CNNs, RNNs, AES, and VAES; these include (i) AML in \cite{Liu2021DeepApproach}; (ii) LAD \cite{Malhotra2015LongSeries}; (iii) LED \cite{Malhotra2016LSTM-basedDetection}; (iv) GE \cite{Guo2018MultidimensionalApproach};
 (v) DIF \cite{Xu2023DeepDetection}; and (vi) TED \cite{Garg2022AnSeries}. 

The evaluation metrics are selected as, Precision, Recall, and F1-score. A threshold $\beta$ or $\tau$ will be employed to maximise the F1-score, such that for the input sequence $\mathbf{x}_\text{test}$, it is classified as an anomaly if $\hat p(h_{\mathbf{x}_\text{test}}^E) < \beta$ (also see Equation \ref{eq:decision_alpha}).
For the benchmark algorithms, anomaly scores $a(\mathbf{x}_i)$ are computed following their original settings (detailed in \cite{Liu2021DeepApproach, Malhotra2015LongSeries, Malhotra2016LSTM-basedDetection, Guo2018MultidimensionalApproach, Xu2023DeepDetection,Garg2022AnSeries}). If $a(\mathbf{x}_{i}) > \tau$, the window contains $\mathbf{x}_{i}$ will be classified as an anomaly. 
All the experiments are repeated ten times, and the mean values of the metrics are reported.

\begin{figure}[!htb]
\centering
\includegraphics[trim = 0mm 0mm 0mm 0mm,clip,width=.92\linewidth]{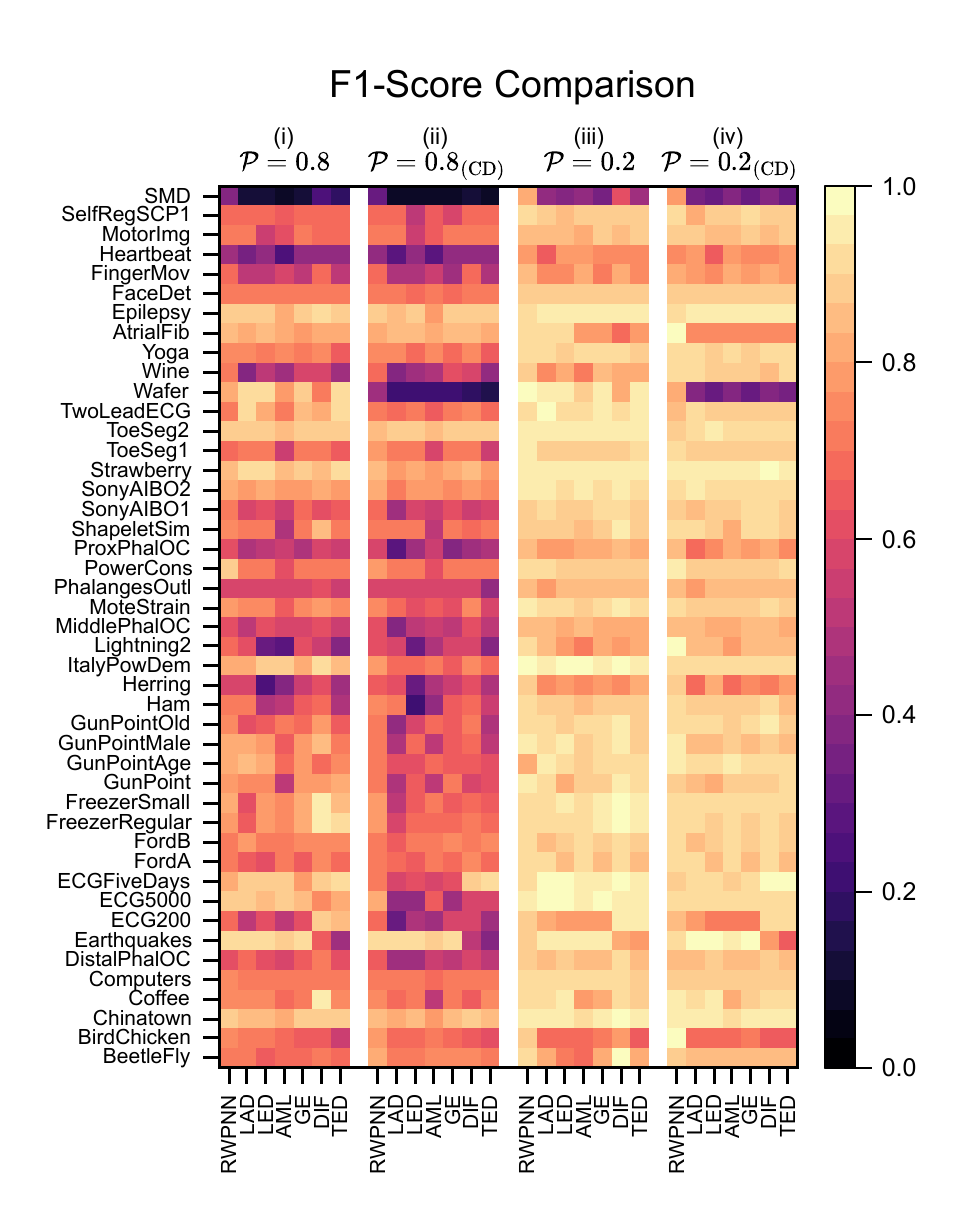}
\caption{
F1-Score performance heatmap for the models and datasets considering $\mathcal{P}=\{0.8, 0.2\}$ and the presence of CD.
The heatmap is segmented into four regions: (i) $\mathcal{P} = 0.8$, (ii) $\mathcal{P} = 0.8$ with CD, (iii) $\mathcal{P} = 0.2$, (iv) $\mathcal{P} = 0.2$ with CD. A warmer colour denotes a higher F1-score while a cooler one indicates a lower F1-score.
} \label{fig:res_0.8}
\end{figure}

The hyperparameters for RWPNN include: three orders of $\phi(x)$, $j_0 = [1,2,3,4,5]$, and $\Gamma = 1/[1, 10, 100, 500, 1000]$ for \textit{MRWPN}; for \textit{SREnc-Dec}, a two layers architecture of the encoder and decoder output feature sizes are searched from the space $[128,64,32,16,8,4,2]$ in descending and ascending order, respectively; the learning rate is determined by using the cyclical learning rate method proposed in \cite{Smith2017CyclicalNetworks} to search the optimal learning rate in the range of [1e-5, 1e-1]; and the maximum training epoch is set to $2000$ with Adam optimiser \cite{Ba2015Adam:Optimization} being used. The batch size is selected within the range of $[1,2,4,8,16,32,64,128,256,512]$, where the maximum size from min({$|\mathcal{D}_{\text{train}}|$} , {$|\mathcal{D}_{\text{v}_1}|$} , {$|\mathcal{D}_{\text{v}_2}|$} , {$|\mathcal{D}_{\text{test}}|$}) is chosen.

\subsection{TSAD Results with $\mathcal{P}=0.8$}\label{sec:res_0.8}

In this section, we present the TSAD results for RWPNN and the benchmark algorithms in univariate and multivariate datasets. We evaluate algorithms against two settings, (i) $\mathcal{P} = \{0.8, 0.2\}$ to examine the performance of the algorithms in data-limited and data-rich scenarios, and (ii) added noise to the test set to mimic the case where concept drift (CD) occurs in the future. 
When $\mathcal{P}=0.8$, the majority of the data is allocated to the test set, whereas at $\mathcal{P}=0.2$, the majority of the data is assigned to the training set. 
Note that we randomly added Gaussian noise with $\mathbf{\mu} = 0.3 * \text{ones}(n)$  and $\Sigma = 0.2 * \text{eye}(n)$ to $30\%$ of $\mathcal{D}_{\text{test}}$. We first show the results when $\mathcal{P} = 0.8$ to simulate a constrained setting when limited training data is available for model training.

\begin{table}[!htp]
\centering
\caption{Average performance comparison with $\mathcal{P} = 0.8$.}
\label{tbl:val_0.8_summary}
\scalebox{0.8}
{
\begin{tabular}{lccc}
\hline \hline
\textit{\textbf{}}    & \textbf{Precision}          & \textbf{Recall}             & \textbf{F1}           \\ \hline
{\textit{RWPNN}}         & $0.6371 \pm 0.13$ & $0.9161 \pm 0.10$ & $\mathbf{0.7400 \pm 0.11}$ \\
\textit{LAD}      & $0.6841 \pm 0.20$ & $0.7778 \pm 0.22$ & $0.6985 \pm 0.16$ \\
\textit{LED} & $0.6632 \pm 0.18$ & $0.7641 \pm 0.22$ & $0.6871 \pm 0.18$ \\
\textit{AML}   & $0.5995 \pm 0.17$ & $0.7591 \pm 0.21$ & $0.6481 \pm 0.17$ \\
\textit{GE}       & $0.6463 \pm 0.17$ & $0.8140 \pm 0.16$ & $0.7037 \pm 0.15$ \\           
\textit{DIF}       & $0.6673 \pm 0.17$ & $0.8665 \pm 0.13$ & $0.7364 \pm 0.14$ \\           
\textit{TED}       & $0.6687 \pm 0.18$ & $0.7664 \pm 0.20$ & $0.6904 \pm 0.17$ \\
\hline
\multicolumn{4}{c}{\textbf{With CD}}                                                                                 \\ \hline
{\textit{RWPNN}}         & $0.5926 \pm 0.12$ & $0.9369 \pm 0.09$ & $\mathbf{0.7167 \pm 0.11}$ \\
\textit{LAD}      & $0.5700 \pm 0.17$ & $0.6994 \pm 0.28$ & $0.5983 \pm 0.18$ \\
\textit{LED} & $0.5791 \pm 0.16$ & $0.6975 \pm 0.25$ & $0.6059 \pm 0.18$ \\
\textit{AML}   & $0.5627 \pm 0.15$ & $0.7426 \pm 0.21$ & $0.6149 \pm 0.16$ \\
\textit{GE}       & $0.5794 \pm 0.16$ & $0.7668 \pm 0.20$ & $0.6410 \pm 0.16$ \\
\textit{DIF}       & $0.5865 \pm 0.15$ & $0.7928 \pm 0.20$ & $0.6604 \pm 0.16$ \\
\textit{TED}       & $0.5772 \pm 0.16$ & $0.6854 \pm 0.24$ & $0.6053 \pm 0.17$ \\ \hline \hline
\end{tabular}
}
\end{table}

The mean metrics with standard deviation are calculated and summarised in Table \ref{tbl:val_0.8_summary}. We highlighted the best F1-score for a better visualisation. 
We further present a detailed F1-Score comparison using a heatmap in Fig. \ref{fig:res_0.8}, where (i) shows the F1-score for each dataset when $\mathcal{P}=0.8$, (ii) is the results for $\mathcal{P}=0.8$ with CD presents, (iii) shows the results of $\mathcal{P}=0.2$, and (iv) is the results $\mathcal{P}=0.2$ with CD. The warmer the colour, the higher the value of the F1-Score.

From Table \ref{tbl:val_0.8_summary} and the heatmap in Fig. \ref{fig:res_0.8} (i), given the scarcity of the training data, RWPNN outperforms all benchmark algorithms in terms of Recall and F1-score. It has a lower Precision value due to $\beta$ being chosen to maximise the F1-score, and therefore, a larger value of $\beta$ is selected by the decision algorithm, which leads to at least 5\% higher Recall value being reported compared to the benchmark algorithms.
Note that some datasets, e.g., \textit{BeetleFly} and \textit{BirdChicken}, report higher performance with CD, this is due to $\mathcal{A}_{\text{perc}}$ is higher than others, and therefore, when randomly adding the drift to the test set, the anomaly class is more likely to be selected, thus increasing the class differences and easing the AD task. 

In the more extreme scenario when the CD is present {(also see Fig. \ref{fig:res_0.8} (ii))}, RWPNN outperforms all the benchmark algorithms in Precision, Recall, and F1-Score. Regarding the performance difference of RWPNN in the case with and without CD, it has only $4\%$ and $2\%$ decreases in Precision and F1-Score, and $2\%$ increases in Recall. Compared to the results from the best benchmark algorithm, RWPNN reports a $6\%$ improvement in the F1-score.

We note that for the SMD dataset, performance is suboptimal across all algorithms. This is due to SMD being the most challenging dataset with $28$ entities and only $4.3\%$ of anomalies present in the dataset. 
For $\mathcal{P}=0.8$, the F1-scores of the benchmark algorithms range between $[0.11, 0.26]$, with the highest at $0.26$ achieved by DIF. In contrast, RWPNN outperforms all benchmarks by reporting at least $13\%$ higher F1-score due to the powerful latent space modelling capability provided by the \textit{MRWPN} module. In the scenario of $\mathcal{P}=0.8$ with the presence of CD, the combined challenges of CD and limited data further complicate the TSAD task. Here, DIF leads the benchmarks with an F1-score of $0.10$. However, RWPNN demonstrates greater robustness to CD, achieving a $22\%$ higher F1-score than the best benchmark result.

Note that existing TSAD studies involving DNNs usually require a large amount of data to train the model; in this paper, we evaluate the models using different $\mathcal{P}$ to simulate the situation with and without sufficient data for the model to learn the task. The proposed RWPNN performs well when the training data is scarce; it outperforms the benchmark algorithms in all the metrics and has less performance degradation when CD exists in the test data.

\subsection{TSAD Results with $\mathcal{P}=0.2$}

In this section, we show the results when $\mathcal{P}=0.2$, where all the models have sufficient data for training and are expected to perform better compared to the previous section, given that all the models will not be constrained by data availability.

Fig. \ref{fig:res_0.8} (iii-iv) shows the F1-score comparison among the seven algorithms under conditions with and without CD, with the aggregated performance detailed in Table \ref{tbl:val_0.2_summary}. We can see that with sufficient data, all models exhibit enhanced performance, leading to a generally warmer colour scheme in Fig. \ref{fig:res_0.8} (iii-iv) as opposed to Fig. \ref{fig:res_0.8} (i-ii). In these conditions, RWPNN consistently outperforms others, demonstrating an improvement of $1\%-5\%$ in Precision, $2\%-7\%$ in Recall, and $2\%-5\%$ in F1-Score, thereby maintaining its superior TSAD performance.

Similar to the discussion with $\mathcal{P}=0.8$, when CD exists, all models tend to degrade their performance; however, RWPNN outperforms all benchmark algorithms in all metrics. It is worth highlighting the performance drop for the dataset \textit{Wafer}, as it is the one with the lowest $\mathcal{A}_{\text{perc}}$, and adding CD to $\mathcal{D}_{\text{test}}$ introduces the greatest difficulty among all the datasets. While all the benchmark algorithms fail to find the optimal decision boundary for TSAD (see Fig. \ref{fig:res_0.8}), RWPNN still maintains its accurate and robust performance against CD.

In the case of the SMD dataset, the availability of more training data leads to improved F1-scores across all models. DIF achieves the best benchmark score of 0.61 for $\mathcal{P}=0.2$, while AML leads with 0.38 for $\mathcal{P}=0.2$ in the presence of CD. RWPNN, benefiting from the proposed \textit{MRWPN} module, still outperforms these benchmarks, demonstrating F1-score enhancements of 22\% and 40\% for the respective scenarios.

\begin{table}[!htp]
\centering
\caption{Average performance comparison with $\mathcal{P} = 0.2$.}
\label{tbl:val_0.2_summary}
\scalebox{0.8}
{
\begin{tabular}{lccc}
\hline \hline
\textit{\textbf{}}    & \textbf{Precision}          & \textbf{Recall}             & \textbf{F1}           \\ \hline
{\textit{RWPNN}}         & $0.8653 \pm 0.08$ & $0.9628 \pm 0.05$ & $\mathbf{0.9063 \pm 0.04}$ \\
\textit{LAD}      & $0.8584 \pm 0.10$ & $0.9016 \pm 0.12$ & $0.8754 \pm 0.10$ \\
\textit{LED} & $0.8458 \pm 0.10$ & $0.8978 \pm 0.11$ & $0.8683 \pm 0.10$ \\
\textit{AML}   & $0.8118 \pm 0.10$ & $0.9183 \pm 0.11$ & $0.8575 \pm 0.10$ \\
\textit{GE}       & $0.8363 \pm 0.11$ & $0.9068 \pm 0.11$ & $0.8676 \pm 0.10$ \\
\textit{DIF}       & $0.8454 \pm 0.10$ & $0.9445 \pm 0.08$ & $0.8892 \pm 0.08$ \\
\textit{TED}       & $0.8450 \pm 0.10$ & $0.8990 \pm 0.10$ & $0.8683 \pm 0.09$ \\
\hline
\multicolumn{4}{c}{\textbf{With CD}}                                                                                 \\ \hline
{\textit{RWPNN}}         & $0.8520 \pm 0.08$ & $0.9652 \pm 0.05$ & $\mathbf{0.9008 \pm 0.05}$ \\
\textit{LAD}      & $0.8311 \pm 0.13$ & $0.8697 \pm 0.15$ & $0.8453 \pm 0.13$ \\
\textit{LED} & $0.8259 \pm 0.13$ & $0.8668 \pm 0.15$ & $0.8425 \pm 0.13$ \\
\textit{AML}   &  $0.8032 \pm 0.13$ & $0.8994 \pm 0.13$ & $0.8436 \pm 0.12$ \\
\textit{GE}       & $0.8197 \pm 0.13$ & $0.8834 \pm 0.14$ & $0.8473 \pm 0.13$ \\
\textit{DIF}       & $0.8197 \pm 0.13$ & $0.9088 \pm 0.13$ & $0.8585 \pm 0.12$ \\
\textit{TED}       & $0.8245 \pm 0.13$ & $0.8765 \pm 0.15$ & $0.8460 \pm 0.13$ \\ \hline \hline
\end{tabular}
}
\end{table}

\subsection{Early Warning Detection}
The proposed RWPNN leverages statistical interpretations of the data distribution for TSAD, which can also be useful when implementing an early warning system for healthcare monitoring. 


Here, we conduct an analysis on the electrocardiogram dataset \textit{ECG5000} from \cite{Anh2018TheArchive}, and show how RWPNN can contribute to an early warning healthcare monitoring system. We adopt the best hyperparameter settings from Section \ref{sec:res_0.8} with $\mathcal{P}=0.8$; and select the reconstruction-based model \textit{LED} for comparison. Note that for the early warning analysis, we train the \textit{MRWPN} with the latent features $y^E_{\mathbf{x}}$ from the last layer of the Encoder (see Fig. \ref{fig:rwpnn_flowchart}) to get $\hat{p}(y^E_{\mathbf{x}})$.

\begin{figure}[!htp]
\centering
\includegraphics[trim = 0mm 0mm 0mm 0mm,clip,width=0.95\linewidth]{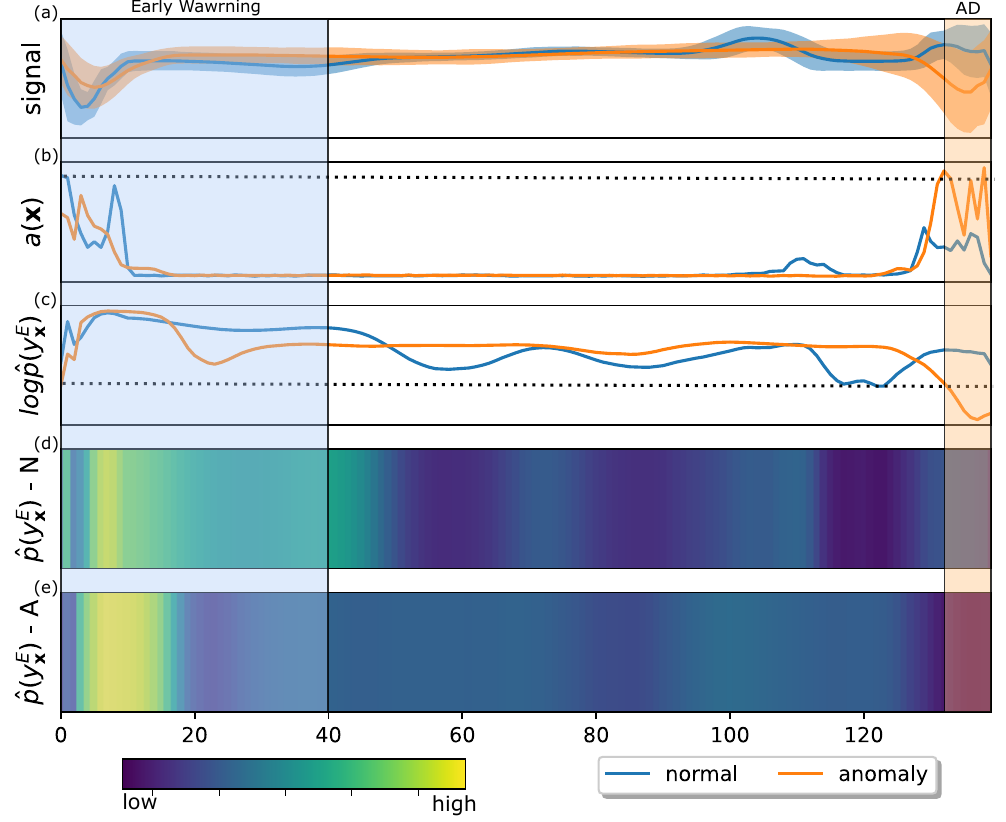}
\caption{
The estimated PDFs for the \textit{ECG5000} dataset are depicted. (a) shows the mean and highlights the standard deviation of the normal and anomaly classes. (b) is the anomaly scores $a(\mathbf{x})$ for \textit{LED}. (c) shows the scores generated $\hat{p}(y^E_{\mathbf{x}})$ by RWPNN. (d-e) are the pattern variations of $\hat{p}(y^E_{\mathbf{x}})$ for the normal and anomaly classes, respectively.
}
\label{fig:pdf_drift}
\end{figure}

Fig. \ref{fig:pdf_drift} illustrates the mean values of the electrocardiograms and two detection metrics using the signal $\mathbf{x}$ with $L=140$: the anomaly scores $a(\mathbf{x})$, and the estimated PDFs for the latent features $\hat{p}(y^E_{\mathbf{x}})$. 
In subfigures (b) and (c), the dotted line represents a potential decision boundary for TSAD in the \textit{LED} and the proposed RWPNN. The subfigures (d) and (e) further illustrate the pattern variations of the estimated PDFs for the normal and anomaly classes, indicating lower values with cooler colours. Additionally, two regions are highlighted in orange and blue, denoted as the TSAD window (on the right of the figure) and the Early Warning window (on the left of the figure), respectively.

As observed in Fig. \ref{fig:pdf_drift}(a) and (b), the peak variations of the signal at $0 \leq t \leq 10$ do not have sufficient information to distinguish differences between normal and anomalous ECGs in the early stage. 
The anomaly becomes identifiable around $t=130$, as shown in Fig. \ref{fig:pdf_drift}(b), where distinct oscillation patterns emerge, which are particularly highlighted in the AD window on the right. 

Furthermore, RWPNN outperforms the \textit{LED} in capturing subtle class differences and anomaly precursors, particularly in regions like $20 \leq t \leq 40$, where both the original signal and anomaly scores exhibit similar patterns (as seen in Fig. \ref{fig:pdf_drift}(a-b)).

\begin{figure}[!htp]
\centering
\includegraphics[trim = 0mm 0mm 0mm 0mm,clip,width=\linewidth]{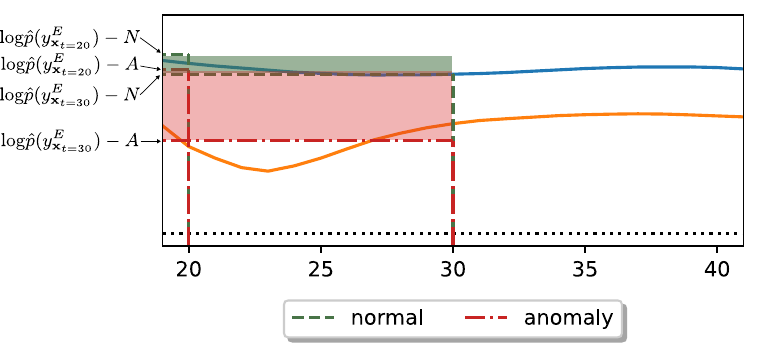}
\caption{
A sliding window-based Early Warning System monitoring the change in $\hat{p}(y^E_{\mathbf{x}_t})$.
}
\label{fig:CYB_early_warning}
\end{figure}

Distinctive pattern variations between the normal and anomaly classes can be observed in Fig. \ref{fig:pdf_drift} (d-e). By monitoring the variations of $\hat{p}(y^E_{\mathbf{x}})$ generated by the RWPNN, anomaly precursors can be identified prior to the actual event. 

To illustrate, we select a timeframe of $20\leq t \leq 40$ and implement a simple rolling window-based Early Warning System to detect the precursors using the generated $\hat{p}(y^E_{\mathbf{x}})$. This system is depicted in Fig. \ref{fig:CYB_early_warning}, where the four labels on the y-axis correspond to the logarithmic values of $\hat{p}(y^E_{\mathbf{x}_t})$ for the normal, expressed as $\hat{p}(y^E_{\mathbf{x}_t}) - N$, and the anomaly data, denoted as $\hat{p}(y^E_{\mathbf{x}_t}) - A$, at $t=$ $20$ and $30$, respectively.

In Fig. \ref{fig:CYB_early_warning}, the green area denotes the difference in the estimated $\hat{p}(y^E_{\mathbf{x}_t}) - N$ for the normal data at $t=$ 20 and 30, whereas the red area indicates the difference for the anomaly data. These differences are calculated by computing the change in $\hat{p}(y^E_{\mathbf{x}_t})$ at $t=20$ and $t=30$, employing a rolling-mean window of size $s=5$ for each computation. We note that the red area emphasises a more significant variation in the estimated $\hat{p}(y^E_{\mathbf{x}_t})$ for the anomaly data within this early stage timeframe. Therefore, an Early Warning System can be created by analysing the $\hat{p}(y^E_{\mathbf{x}})$. Specifically, when the deviation of $\hat{p}(y^E_{\mathbf{x}_t})$ between two timestamps exceeds a predefined threshold, a precursor alert is triggered in the system.

\subsection{Ablation Test}

Previous sections have discussed the robust detection performance of RWPNN in the non-stationary environment with different data availability cases. This has highly relied on the \textit{MRWPN}, so we further conduct an ablation test on \textit{MRWPN} by replacing it with other traditional AD algorithms, i.e., One-Class Support Vector Machine (OCSVM), Local Outlier Factor (LOF), Isolation Forest (IForest) \cite{Scholkopf2001EstimatingDistribution, Breunig2000LOF:Outliers, Liu2008IsolationForest}, using \textit{scikit-learn} \cite{scikit-learn}
, 
and with state-of-the-art AD algorithms, i.e., DIF, GOAD, DSVDD \cite{Xu2023DeepDetection, Bergman2020Classification-BasedData, Ruff2018DeepClassification}.
Note that to assess the performance of the MRWPN module across varying levels of data scarcity, the ablation tests are also conducted using $\mathcal{P}=\{0.8,0.2\}$.

For each of the traditional methods, we selected the best configuration from (i) kernels \textit{[linear, poly, rbf, sigmoid]}, gamma values \textit{[auto, scale]}; (ii) two neighbours; and (iii) estimator number of \textit{[10, 50, 100]}, for OCSVM, LOF, and IForest, respectively. For DIF, GOAD, and DSVDD, we follow the default settings in the respective papers. 

The mean results of each metric are reported in Table \ref{tbl:ablation}. The results demonstrate the importance of the proposed \textit{MRWPN} module, as other methods modelling the latent space fail to provide reliable results in all scenarios.

\begin{table}[!htp]
\centering
\caption{Ablation test using $\mathcal{P}=\{0.8,0.2\}$ with and without CD.}
\label{tbl:ablation}
\scalebox{0.8}
{
\begin{tabular}{lccc}
\hline \hline
                 & \textbf{Precision}          & \textbf{Recall}             & \textbf{F1}                 \\ \hline
\multicolumn{4}{c}{\textbf{$\mathcal{P}=0.8$ w/o CD}}                                                      \\
{\textit{RWPNN}}    & $0.6371 \pm 0.13$ & $0.9161 \pm 0.10$ & $\mathbf{0.7400 \pm 0.11}$ \\
\textit{OCSVM}   & $0.5538 \pm 0.18$ & $0.6759 \pm 0.23$ & $0.5858 \pm 0.19$ \\
\textit{LOF}     & $0.4939 \pm 0.18$ & $0.6343 \pm 0.26$ & $0.5251 \pm 0.21$ \\
\textit{IForest} & $0.4179 \pm 0.22$ & $0.4373 \pm 0.30$ & $0.3996 \pm 0.25$ \\
\textit{DIF} & $0.5589 \pm 0.13$ & $0.8356 \pm 0.20$ & $0.6470 \pm 0.14$ \\
\textit{GOAD} & $0.5485 \pm 0.12$ & $0.6071 \pm 0.21$ & $0.6333 \pm 0.18$ \\
\textit{DSVDD} & $0.5486 \pm 0.13$ & $0.8418 \pm 0.20$ & $0.6431 \pm 0.16$ \\
\multicolumn{4}{c}{\textbf{$\mathcal{P}=0.8$ w/ CD}}                                                       \\
{\textit{RWPNN}}    & $0.5926 \pm 0.12$ & $0.9369 \pm 0.09$ & $\mathbf{0.7167 \pm 0.11}$ \\
\textit{OCSVM}   & $0.5552 \pm 0.15$ & $0.7285 \pm 0.19$ & $0.6078 \pm 0.16$ \\
\textit{LOF}     & $0.5182 \pm 0.20$ & $0.5882 \pm 0.31$ & $0.5028 \pm 0.24$\\
\textit{IForest} & $0.4850 \pm 0.21$ & $0.5152 \pm 0.31$ & $0.4599 \pm 0.25$ \\
\textit{DIF} & $0.5507 \pm 0.13$ & $0.8442 \pm 0.16$ & $0.6507 \pm 0.14$ \\
\textit{GOAD} & $0.5553 \pm 0.16$ & $0.8429 \pm 0.21$ & $0.6503 \pm 0.18$ \\
\textit{DSVDD} & $0.5482 \pm 0.14$ & $0.8711 \pm 0.19$ & $0.6578 \pm 0.16$ \\
\multicolumn{4}{c}{\textbf{$\mathcal{P}=0.2$ w/o CD}}                                                      \\
{\textit{RWPNN}}    & $0.8653 \pm 0.08$ & $0.9628 \pm 0.05$ & $\mathbf{0.9063 \pm 0.04}$ \\
\textit{OCSVM}   & $0.5883 \pm 0.23$ & $0.6244 \pm 0.18$ & $0.7695 \pm 0.17$ \\
\textit{LOF}     & $0.7694 \pm 0.15$ & $0.5984 \pm 0.25$ & $0.6321 \pm 0.21$ \\
\textit{IForest} & $0.6517 \pm 0.29$ & $0.3511 \pm 0.24$ & $0.4273 \pm 0.25$ \\
\textit{DIF} & $0.7785 \pm 0.11$ & $0.9376 \pm 0.17$ & $0.8410 \pm 0.15$ \\
\textit{GOAD} & $0.7873 \pm 0.13$ & $0.9522 \pm 0.14$ & $0.8523 \pm 0.12$ \\
\textit{DSVDD} & $0.7889 \pm 0.11$ & $0.9554 \pm 0.10$ & $0.8563 \pm 0.11$ \\
\multicolumn{4}{c}{\textbf{$\mathcal{P}=0.2$ w/ CD}}                                                       \\
{\textit{RWPNN}}    &  $0.8520 \pm 0.08$ & $0.9652 \pm 0.05$ & $\mathbf{0.9008 \pm 0.05}$ \\
\textit{OCSVM}   & $0.7938 \pm 0.18$ & $0.5664 \pm 0.22$ & $0.6318 \pm 0.20$ \\
\textit{LOF}     & $0.7254 \pm 0.22$ & $0.5255 \pm 0.31$ & $0.5559 \pm 0.28$ \\
\textit{IForest} & $0.6300 \pm 0.28$ & $0.3697 \pm 0.26$ & $0.4341 \pm 0.27$ \\
\textit{DIF} & $0.7677 \pm 0.17$ & $0.9140 \pm 0.19$ & $0.8279 \pm 0.18$ \\
\textit{GOAD} & $0.7995 \pm 0.11$ & $0.9736 \pm 0.04$ & $0.8727 \pm 0.09$ \\
\textit{DSVDD} & $0.7672 \pm 0.17$ & $0.9136 \pm 0.19$ & $0.8276 \pm 0.18$ \\
\hline \hline
\end{tabular}
}
\end{table}

One key contribution of the proposed RWPNN is its \textit{MRWPN} module, which achieves lower time complexity than the WPNN proposed in \cite{Garcia-Trevino2024WaveletNetworks} by computing multiple receptive fields in parallel. 

In the remaining section, we compare the F1-score and computation time of these modules by measuring the time for training and evaluating the modules, which involve the coefficient update and density estimation steps. Here, we run the test using the seven UEA datasets used in Fig. \ref{fig:res_0.8} and report the F1-Score in Table \ref{tbl:CYB_ablation_MRWPN_f1}, for $\mathcal{P}=0.8$.

Regarding the computation time, we note that the multi-receptive field approach of \textit{MRWPN} eliminates the need to construct separate WPNNs, thus streamlining the overall training and testing durations. For instance, \textit{WPNN} requires 11.7 seconds to train and test five different receptive fields, while \textit{MRWPN} completes in only 7.3 seconds. Furthermore, by considering results from all receptive fields in parallel, \textit{MRWPN} reports an F1-score at least $2\%$ higher than the best WPNN with $\alpha=1/1000$.

\begin{table}[!htp]
\centering
\caption{Ablation test for the proposed MRWPN.}
\label{tbl:CYB_ablation_MRWPN_f1}
\scalebox{0.8}
{
\begin{tabular}{lc}
\hline\hline
\multicolumn{1}{c}{\textbf{Model}} & \multicolumn{1}{c}{\textbf{F1-Score    }} \\ \hline 
MRWPN                              & $\mathbf{0.6972\pm0.02}$                                \\
WPNN $\alpha=1$                    & $0.5192\pm0.14$                                \\
WPNN $\alpha=1/10$                 & $0.6751\pm0.02$                                \\
WPNN $\alpha=1/100$                & $0.6752\pm0.02$                                \\
WPNN $\alpha=1/500$                & $0.6764\pm0.02$                                \\
WPNN $\alpha=1/1000$               & $0.6779\pm0.02$     \\                          \hline\hline
\end{tabular}
}
\end{table}

\subsection{Summary}

The experiments with different data availability and the presence of CD highlight the fact that: the fewer data samples are available for training, the harder it is for the model to tolerate variations in the dataset. The proposed RWPNN demonstrates outstanding TSAD performance, especially with ECG, EEG, and sensor measurements, where issues like noise perturbations, training data scarcity, and data non-stationary issues are more likely to occur. This benefit comes from the robust latent space modelling capability provided by \textit{SREnc-Dec} and \textit{MRWPN}. However, the RWPNN reports lower performance when the dataset contains more spatial and global features, as the \textit{SREnc-Dec} is only based on the RNNs. This weakness can be effectively addressed in future works by employing CNNs in the feature extraction stage.

Fig. \ref{fig:res_comparison} shows the F1-Score performance comparison for tasks with $\mathcal{P}=\{0.8,0.2\}$, both with and without CD, where RWPNN is benchmarked against other models. DIF exhibits the least performance variation across different values of $\mathcal{P}$. This stability is attributed to the use of a randomly initialised encoder in DIF, which facilitates diverse feature representations, thereby enhancing robustness against CD. In contrast, RWPNN outperforms all methods, attributed to its more effective and robust density estimation approach, leading to superior performance in all experimental scenarios.

{\color{red}
\begin{figure}[!htp]
\centering
\includegraphics[trim = 0mm 0mm 0mm 0mm,clip,width=0.9\linewidth]{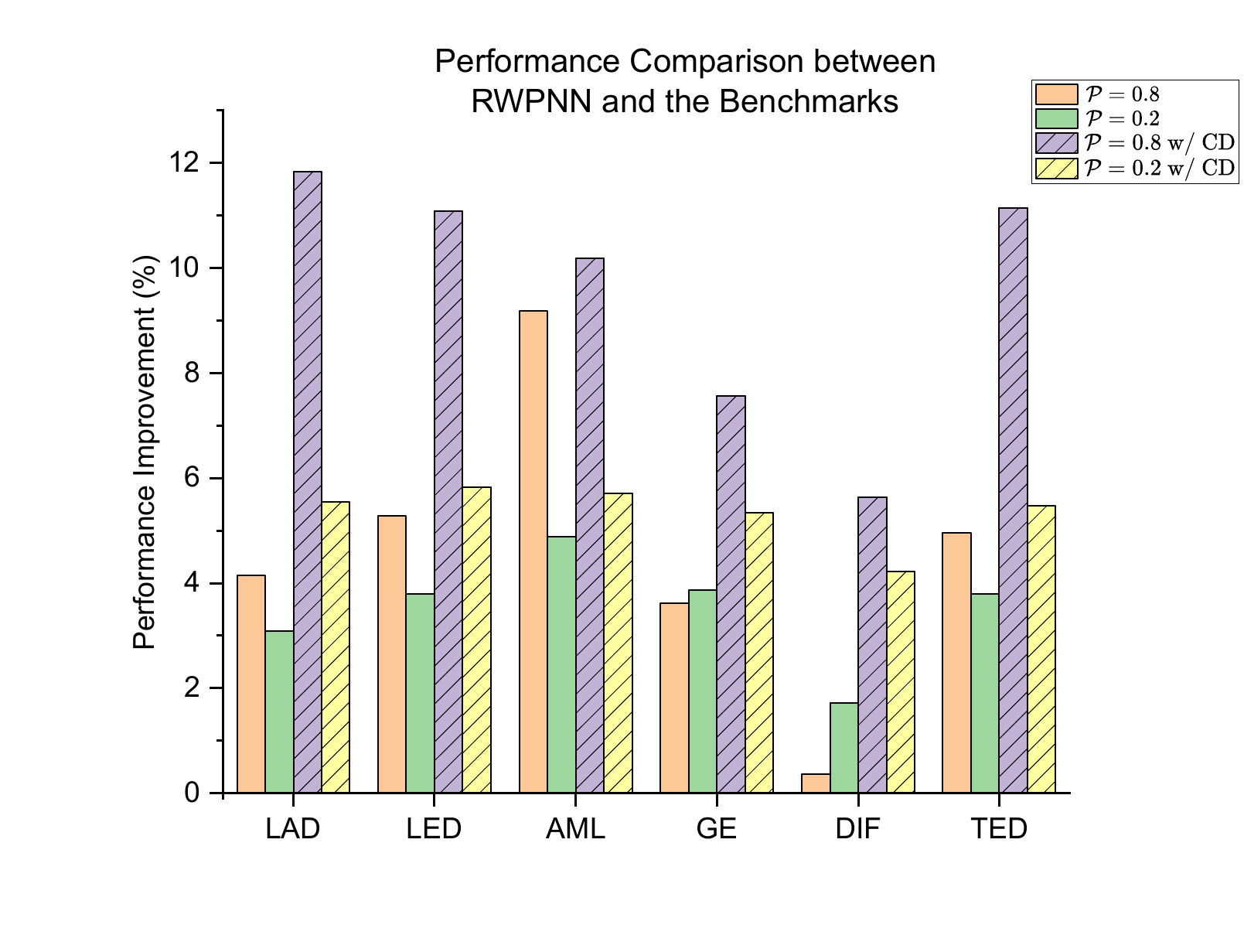}
\caption{
Summary of the performance difference between RWPNN and the benchmark algorithms measured by F1-score.
} \label{fig:res_comparison}
\end{figure}
}

\clearpage
\section{Conclusion}\label{sec:final_remark}
In this paper, a novel recurrent wavelet probabilistic neural network is proposed, which is suitable for detecting time series anomalies in a non-stationary environment with limited training data. RWPNN consists of two modules: a \textit{SREnc-Dec} module that extracts temporal features from the time series data, and an ensemble \textit{MRWPN} module that models the probability density functions of the latent space in a nonparametric way.

We extend the application of the current generative-based wavelet probabilistic network from standard neural networks to deep neural networks by combining the \textit{SREnc-Dec} with the \textit{MRWPN}. Meaningful and compressed temporal features are learned in \textit{SREnc-Dec}, which also neutralises the curse of dimensionality that hinders further research on wavelet probabilistic networks. The design of the ensemble \textit{MRWPN} can handle different rates of data variation in non-stationary environments and also removes one step of the hyperparameter tuning process in the existing solution without formulating extra networks. Such ensemble views are further optimised and selected automatically for TSAD tasks. Therefore, RWPNN has strengthened anomaly detection performance compared to the existing wavelet probabilistic network, and the TSAD solutions that rely on the reconstruction or prediction process and modelling the latent space using the parametric density estimation technique.

The framework has been evaluated on $45$ datasets from different domains covering univariate and multivariate data; it has encouraging detection performance in handling imbalanced and scarce time series data with suspected concept drift; superior performance is reported. RWPNN can react quickly to anomalies and can capture subtle class differences by utilising the extracted statistical information from the latent space. The ablation test demonstrates the importance of the \textit{MRWPN} module, which has a better generalisation ability in the non-stationary environment with potential data shortage issues.

The current version of RWPNN only considers the temporal features using the \textit{SREnc-Dec} module. Recent literature has suggested including both temporal and spatial features at the same time. 
Therefore, it is worth developing an extension to RWPNN using 
more DNN architectures, such as CNNs and Transformers, to extract more expressive features. 
Further study can also focus on the early warning system by leveraging different configurations of neural network architectures with \textit{MRWPN}. 
Additionally, developing an adaptive algorithm to automatically select the optimal values for anomaly detection and effective ensemble views can enhance the performance of the proposed RWPNN.

\newpage

\vfill

\bibliographystyle{elsarticle-num}
\bibliography{main}


\end{document}